\documentclass[final]{cvpr}

\usepackage{times}
\usepackage{epsfig}
\usepackage{graphicx}
\usepackage{amsmath}
\usepackage{amssymb}
\usepackage{bbm}

\usepackage[square,numbers]{natbib}
\usepackage[utf8]{inputenc} 
\usepackage[T1]{fontenc}    
\usepackage{url}            
\usepackage{booktabs}       
\usepackage{amsfonts}       
\usepackage{amsmath}
\usepackage{nicefrac}       
\usepackage{microtype}      
\usepackage{graphicx}
\usepackage{xcolor}
\usepackage{enumitem}
\usepackage{subcaption}
\usepackage{multirow}
\usepackage{wrapfig}

\usepackage[pagebackref=true,breaklinks=true,colorlinks,bookmarks=false]{hyperref}

\def\cvprPaperID{738} 
\def\confYear{CVPR 2021}

\title{UV-Net: Learning from Boundary Representations}

\author{%
Pradeep Kumar Jayaraman\\
Autodesk Research
\and
Aditya Sanghi\\
Autodesk Research
\and
Joseph G. Lambourne\\
Autodesk Research
\and
Karl D.D. Willis\\
Autodesk Research
\and
Thomas Davies\\
Autodesk
\and
Hooman Shayani\\
Autodesk Research
\and
Nigel Morris\\
Autodesk Research
}

\newcommand{\bez}{B\'ezier~}
\newcommand{\umin}{u_{\text{min}}}
\newcommand{\umax}{u_{\text{max}}}
\newcommand{\vmin}{v_{\text{min}}}
\newcommand{\vmax}{v_{\text{max}}}
\newcommand{\curv}{\mathbf{C}}
\newcommand{\surf}{\mathbf{S}}

\newcommand{\myparagraph}[1]{\medskip\noindent\textbf{#1~}}

\iftoggle{cvprfinal}{
}

\begin{document}

\maketitle
\iftoggle{cvprfinal}{
}

\begin{abstract}
We introduce UV-Net, a novel neural network architecture and representation designed to operate directly on Boundary representation (B-rep) data from 3D CAD models. The B-rep format is widely used in the design, simulation and manufacturing industries to enable sophisticated and precise CAD modeling operations. However, B-rep data presents some unique challenges when used with modern machine learning due to the complexity of the data structure and its support for both continuous non-Euclidean geometric entities and discrete topological entities.
In this paper, we propose a unified representation for B-rep data that exploits the U and V parameter domain of curves and surfaces to model geometry, and an adjacency graph to explicitly model topology. This leads to a unique and efficient network architecture, UV-Net, that couples image and graph convolutional neural networks in a compute and memory-efficient manner. To aid in future research we present a synthetic labelled B-rep dataset, SolidLetters, derived from human designed fonts with variations in both geometry and topology. Finally we demonstrate that UV-Net can generalize to supervised and unsupervised tasks on five datasets, while outperforming alternate 3D shape representations such as point clouds, voxels, and meshes.
\end{abstract}

\section{Introduction}
Parametric curves and surfaces form the basis of computer-aided design (CAD) and are widely used in design, simulation, and manufacturing.
CAD software is primarily concerned with modeling and representing 3D solids---closed, watertight shapes which describe objects unambiguously with consistently oriented patches of surface geometry.
The industry-wide standard to represent solid models is the Boundary representation (B-rep)~\cite{weiler1986:topological,lee2001:partialentity}. The B-rep is a versatile data structure comprised of faces (bounded portions of surfaces), edges (bounded pieces of curves) and vertices (points), glued together with topological connections between them.
The B-rep enables a variety of parametric curves and surfaces, such as lines, arcs, planes, cylinders, toruses and Non-Uniform Rational B-Splines (NURBS), to precisely represent complex 3D shapes formed from CAD modeling operations such as extrusions, fillets, and Booleans.
CAD users interact directly with B-rep faces, edges, and vertices to select, align, and modify 3D shapes. To leverage the recent advances of deep neural networks in CAD software, an appropriate representation of B-rep data is required. Such a representation has the potential to unlock numerous CAD applications such as auto-complete of modeling operations, smart selection tools, shape similarity search and many more. Critical to enabling these applications is a representation that encodes the B-rep entities themselves.
\begin{figure}
    \centering
    \includegraphics[width=0.98\columnwidth]{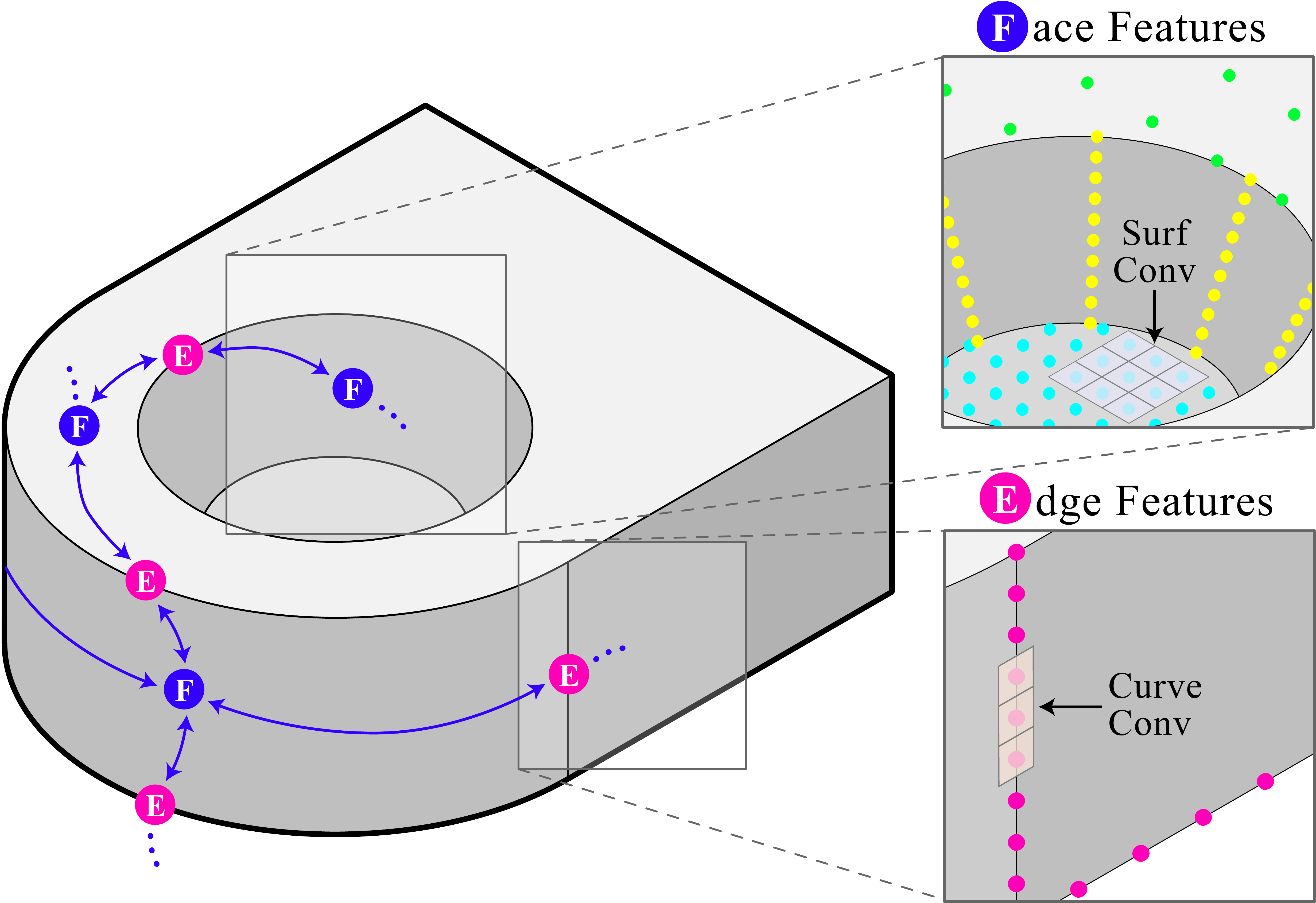}
    \caption{UV-Net builds features by sampling points on the edges and faces of solid models. These features are then message-passed among adjacent topological entities.}
    \label{fig:teaser}
\end{figure}

Despite widespread usage of B-rep data in the industry, there exists limited research on applying deep neural networks to this representation directly.
There are numerous challenges in feeding B-rep data to neural networks.
B-rep data consists of disparate geometric and topological entities, such as parametric curves and surfaces, each with their own set of parameters. 
Moreover, the mapping between a shape and a surface type is not one-to-one, for example, a plane can be represented as a B-spline of arbitrary degree. This means raw surface information, such as parametric coefficients or spline control points, cannot be fed directly into a neural network, as it would not be invariant to the specific parameterization.
Finally, consideration must be given to how different curve and surface geometry are connected to form the entire shape, i.e. the topology.

An alternate approach is to preprocess B-rep data into well-studied representations, such as images, voxels, point clouds, or triangle meshes.
Although plausible, such conversions are neither differentiable, nor trivial. Discretized representations, such as point clouds or voxels, suffer from loss of fidelity and may lose the critical mapping back to the original B-rep entities. 
Conversion to triangle meshes can be non-trivial and prone to failure when high quality, manifold meshes are required~\cite{christophe2009:gmsh}.

To tackle these challenges, we present UV-Net, a novel neural network architecture and representation designed to operate directly on B-rep entities (Figure~\ref{fig:teaser}). In this paper, we make the following contributions:
\begin{itemize}[leftmargin=*]
	\item We present a new representation of 3D CAD models derived from B-reps, which captures geometric features from the parameter domain as a regular grid, and topological information as a graph.
	\item We propose a novel architecture which couples an image CNN and a hierarchical graph-neural network in a compute and memory-efficient manner.
    \item We create and release a synthetic labeled dataset: SolidLetters, which unlike other synthetic datasets, is balanced, and has variations in both geometry and topology.
    \item We demonstrate the efficacy of UV-Net on multiple tasks including 3D shape classification, segmentation, and self-supervised shape retrieval on unlabeled data. We achieve state-of-the-art results on both classification and segmentation tasks by leveraging the full B-rep data structure.
\end{itemize}

\section{Related Work}
\myparagraph{Common geometric representations}
Modern neural networks work with several discrete 3D representations like point clouds, voxels, meshes and multi-view images.
A B-rep can be easily sampled to obtain point clouds~\cite{qi2017:pointnet,wang2019:dgcnn}.
A problem with this conversion is that CAD models often contain small features that convey important information.
A prohibitively dense point cloud might be required to capture such fine details.
3D CNNs~\cite{wu2015:3dShapeNets} can be applied to voxelized B-reps, as shown by Zhang et al.~\cite{zhang2018:featurenet} for classification.
Unfortunately, there is a cubic compute and memory cost to increasing the voxel grid resolution to capture small faces from a B-rep.
O-CNN~\cite{wang2017:ocnn} can alleviate this problem using sparse octrees, however, very deep octrees may be needed to delineate tiny faces common in B-rep data.
Neural networks representing signed distance~\cite{park2019:deepsdf,chen2019:imnet} or occupancy~\cite{mescheder2019:occupancynet} functions are grid-free and concise, but need to learn mappings to B-rep face and edges along with positional encodings to support downstream applications, which is a challenging problem.
Triangle meshes on the other hand better preserve the geometric and topological information of a solid model~\cite{hanocka2019:meshcnn,wiersma2020:surfrotequivcnn}.
These methods require the B-rep to be converted to watertight, manifold meshes with tight constraints on vertex/edge count, edge length, and angles; a difficult task prone to failure~\cite{hu2019:triwild,christophe2009:gmsh}.
Finally, multi-view images represent 3D shapes by rendering, and have shown excellent results on shape classification and retrieval~\cite{su2015:mvcnn}.
Such renderings are not expressive enough to represent and map back to the multitude of entities in a B-rep, thereby limiting applications.
There is a recent interest in the machine learning community
in the generation of parametric geometry such as \bez curves \cite{lopes2019:svgvae,wang2020:attr2font,smirnov2020:dps}, splines \cite{gao2019:deepspline}, Coons patches \cite{smirnov2020:coons} and binary space partitioning planes \cite{chen2020:bspnet}.
However, these methods do not deal with feature extraction from B-reps with disparate parametric curve/surface types.

\myparagraph{3D geometry as images}
Closely related to our consideration of geometry as regular grids or images are geometry images~\cite{gu2002:geometryimages} where arbitrary meshes were parametrized into 2D grids for compression and resampling.
Sinha et al.~\cite{sinha2016:learngeomimages} parametrized meshes globally as images to apply CNNs.
Groueix et al.~\cite{groueix2018:atlasnet} learnt the parametrization of point clouds by deforming grids of points.
Kawasaki et al.~\cite{kawasaki2018:imageprocbspline} smoothed the normal map of  B-spline surfaces for fairing.
Different from these works, we deal with parametric shapes employed in CAD, and focus on deriving a representation from the B-rep while retaining geometric and topological information.

\begin{figure*}
    \centering
    \includegraphics[width=0.92\linewidth]{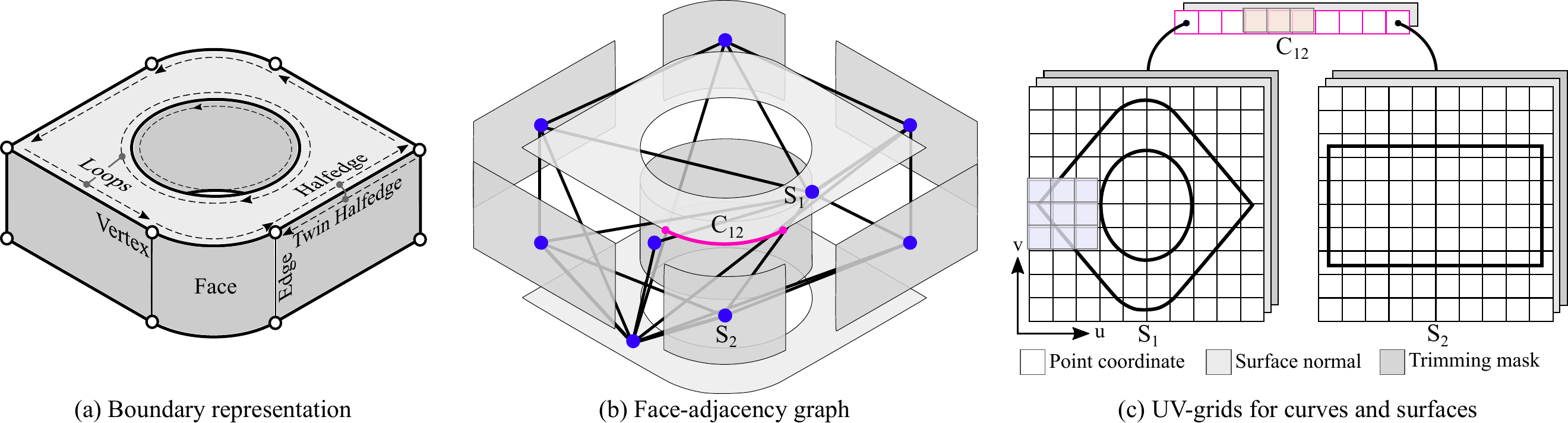}
    \caption{Our representation. (a) The B-rep is a complex data structure with several geometric and topological entities that is difficult to feed to neural networks. (b) We derive a face-adjacency graph from the B-rep to capture topological information. (c) Each face and edge in a B-rep contains a parametric surface and curve, respectively, which we represent as regular UV-grids, and store as node and edge attributes in the graph. Local neighborhoods in UV-grids map to local regions in the geometry.}\label{fig:uvnet_representation}
\end{figure*}

\myparagraph{Boundary representations}
Few neural networks are capable of directly consuming B-rep data.
Initial attempts before the deep learning era focused on automatically recognizing machining features in a solid model.
These methods convert the B-rep into a face-adjacency graph~\cite{ansaldi1985:geometric}, with features such as surface type and edge convexity, that is then used by rule-based schemes~\cite{joshi1988:graphbasedheuristics} or simple neural networks~\cite{prabhakar1992:affm,konstantinos1997:recog212d}.
However, hand-engineered features struggle to generalize well to other tasks.
Babic et al.~\cite{babic2008:reviewrulebased} surveys several classic machine learning methods for feature classification on B-reps.
Very recently, Cao et al.~\cite{cao2020:gnn} used a graph neural network to segment the faces of a B-rep by converting it into a face-adjacency graph.
A major limitation of this method is that it can only work on B-reps with planar faces, as it uses the coefficients of the scalar plane equation as node features to describe the geometry.
In contrast to these works, we aim to derive a general representation from the B-rep that is suitable for a wide range of tasks, and can leverage advances in modern deep learning methods.
\section{Method}
In this section we review the B-rep data structure, and introduce UV-Net's representation and network architecture.
\subsection{Input representation}
\label{sec:input_representation}
The B-rep data structure comprises several topological entities---faces, edges, halfedges, and vertices, with connections between them, see Figure~\ref{fig:uvnet_representation} (a).  Faces are the visible portion of parametric surfaces such as planes, cones, cylinders, toruses, and splines. Edges are the visible interval of parametric curves and vertices are the endpoints of edges.  Each face is delimited by one or more loops of halfedges.  Anti-clockwise loops define outer boundaries while clockwise loops define internal holes.  Solid modeling packages are designed to generate closed and watertight B-rep models, in which every edge contains two halfedges on adjacent faces.  The data structure also stores many references allowing efficient navigation between all adjacent entities \cite{ lee2001:partialentity}. 

Although expressive, the B-rep is a complex data structure and is difficult to feed to neural networks in its original form.
Our goal is to extract the most informative geometric and topological information from the B-rep, and convert it into a representation that can easily and efficiently work with existing neural network architectures.

\myparagraph{Topology}
UV-Net uses a face-adjacency graph derived from the B-rep, $G(V, E)$, to model the topology where the vertices $V$ represent the faces in the B-rep, while the edges $E$ encode the connectivity between the faces, as shown in Figure~\ref{fig:uvnet_representation} (b).
This can be easily built in constant time complexity by traversing through the halfedges of the B-rep: current face $\rightarrow$ halfedges $\rightarrow$ twin-halfedges $\rightarrow$ neighboring faces.
The face adjacency captures the two most geometrically and topologically rich entities: faces and edges from the B-rep, and is sufficient to capture both local and global information about a solid, as we later demonstrate.

\begin{figure*}
    \centering
    \includegraphics[width=0.92\textwidth]{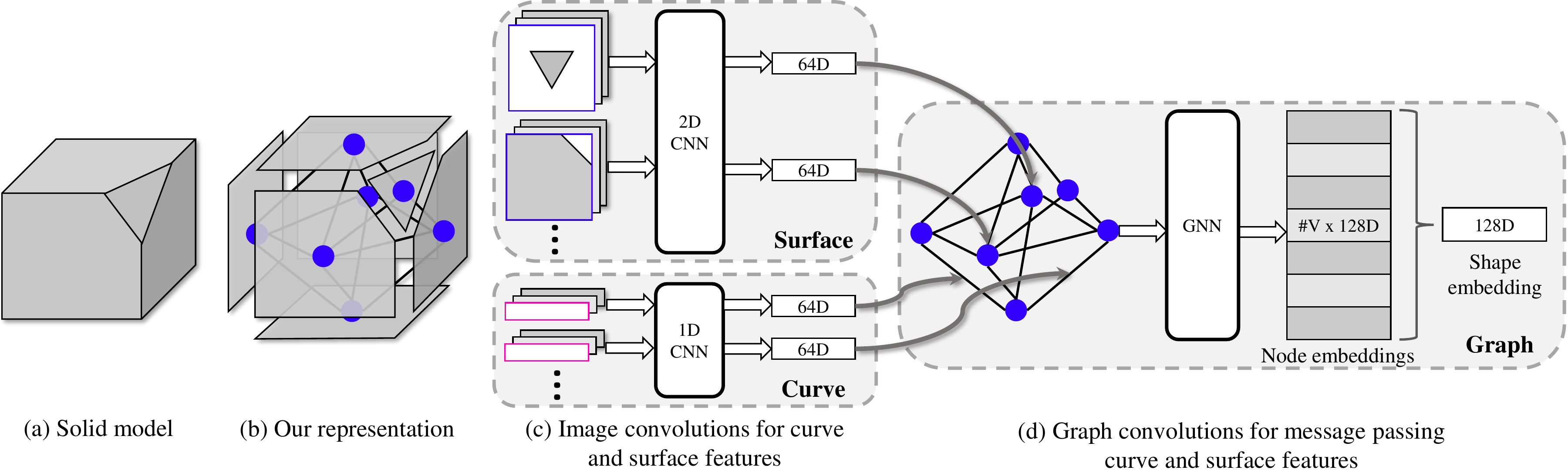}
    \caption{UV-Net encoder architecture. (a) A solid model is represented by (b) a set of regular UV-grids representing each face's and edge's geometry by discretizing the parameter domain, and a graph that captures its topology with face-adjacency information. (c) Curve and surface features are extracted from the UV-grids with 1D and 2D CNNs, respectively. (d) These features are treated as edge and node embeddings of the graph and further processed by graph convolutions. The result is a set of node embeddings, that can be pooled to get the shape embedding of the solid model.}
    \label{fig:network_arch}
\end{figure*}
\myparagraph{Curve geometry}
Each topological edge in a B-rep has an associated parametric curve to define the actual geometry.
Consider one such parametric curve $\curv(u)$, which is a map from an interval $[\umin, \umax] \in \mathbb{R}$, the parameter domain, to the geometry domain $\mathbb{R}^3$.
The curve could be parameterized as a line, circular arc, or B-spline;
we only expect that an interface is available to evaluate the curve and optionally, its first order derivative.
Our idea is to represent the geometry of the curve by discretizing its parameter domain~\cite{kawasaki2018:imageprocbspline} as a regular 1D grid by a uniform step size $\delta u = \frac{\umax - \umin}{M - 1}$, where $M$ is the number of chosen samples, as shown in Figure~\ref{fig:uvnet_representation} (c).
At each of the discretized points in the parameter domain $u_k$, we can attach a set of features evaluated from the curve, e.g.,
absolute point coordinates $\curv(u_k)$, and optionally the unit tangent vector $\hat{\curv_u}(u_k)$ as features.
This 1D UV-grid is set as input edge features in $G$ as shown in Figure~\ref{fig:uvnet_representation}(c).

\myparagraph{Surface geometry}
Each topological face in a B-rep has an associated surface geometry that can be a plane, sphere, cylinder, cone, or a freeform NURBS surface.
The surfaces are trimmed by the halfedge loops that run along the boundary of the face to expose only a portion of the surface as a visible region.
Consider one such parametric surface $\mathbf{S}(u, v)$ which is a map from a 2D interval $[\umin,\umax]\times [\vmin,\vmax] \in \mathbb{R}^2$, the parameter domain, to the geometry domain $\mathbb{R}^3$.
We discretize the parameter domain into a regular 2D grid of samples with step sizes $\delta u = \frac{\umax - \umin}{M - 1}$, and $\delta v = \frac{\vmax - \vmin}{N - 1}$, where $M$ and $N$ are the number of samples along each dimension, as shown in Figure~\ref{fig:uvnet_representation} (c).
The intervals $[\umin, \umax]$ and $[\vmin, \vmax]$ are chosen such that they closely bound the loop that defines the visible region.
At each of these grid points indexed by $(k, l)$, we attach the following local features encoding the geometry of the surface as channels:
(1) 3D absolute point position $\surf(u_k, v_l)$ (the scale of the solid is normalized into a cube of size 2 and centered at origin).~
(2) Optionally, the 3D absolute surface normal $\frac{\surf_u(u_k, v_l)\times \surf_v(u_k, v_l)}{\|\surf_u(u_k, v_l)\times \surf_v(u_k, v_l)\|}$ pointing outwards consistently.
(3) Trimming mask with $1$ and $0$ representing samples that are in the visible region and trimmed region, respectively.
This 2D UV-grid is defined as input node features in $G$.
The representation is flexible enough to incorporate other features like curvature based on the downstream task.

We set the number of samples $M$=$N$=$10$ in all experiments throughout the paper.
This is not a technical restriction, rather it is convenient to form mini-batches of features.
A fixed step size is sufficient when the mapping between parameter and geometry domains are roughly uniform.
We quantitatively evaluate this in the supplementary material.
In the case of extreme parameterizations with high stretching, it is possible to derive a step size to upper bound the distance between samples~\cite{zheng2000:estimatetesselation}.

\myparagraph{Advantages}
The UV-Net representation has several advantages:
(1) Evaluating curves/surfaces at a set of parameters is fast for both primitive and spline surfaces~\cite{sederberg2012:cagd}.
(2) The representation is sparse and scales with the number curves and surfaces in the B-rep.
(3) The grid is largely invariant to the exact parametrization.
For example, the grid does not change when a planar surface is converted into a NURBS patch, or when degree elevation or knot insertion is performed, since the parameterization and geometry remain identical~\cite{piegl1996:nurbsbook}.
In contrast, the raw curve/surface equation will change significantly.
(4) Finally, local neighborhoods in the parameter domain (UV-grids) correspond to local neighborhoods in curve/surface geometry domain, hence hierarchical feature extraction on the manifold~\cite{bronstein2017:geometricdl} is possible.

\subsection{Network architecture}
\label{sec:network_arch}
With this representation, we first perform image convolutions on the curve and surface UV-grids.
These local curve/surface features are then propagated over the entire B-rep with graph convolutions as shown in Figure~\ref{fig:network_arch}.

\myparagraph{Curve \& surface convolution}
Our surface CNN takes in 2D UV-grids with typically 4 or 7 channels (3 xyz, 3 normals, 1 trimming mask) and is defined as:
\(\text{Conv}(4/7, 64, 3)\rightarrow \text{Conv}(64, 128, 3)\rightarrow \text{Conv}(128, 256, 3)
\rightarrow \text{Pool}(1, 1) \rightarrow \text{FC}(256, 64),\)
where \(\text{Conv}(i, o, k)\) is an image convolutional layer with $i$ input channels, $o$ output channels, and kernel size $k$, \(\text{Pool}(n, n)\) is an adaptive average pooling layer which outputs a \(n\times n\) feature map, and \(\text{FC}(i, o)\) is a fully connected layer which takes an input in $i$-D vector and maps it to $o$-D vector.
Our curve CNN takes 1D UV-grids computed from the curves lying in the edges of the B-rep, and is defined similarly with 1D convolutional and pooling layers.
The weights of the curve and surface CNN are shared among all edges and faces in a B-rep, respectively, making them permutation-invariant.
Both convolutional and fully-connected layers do not have biases, and include batch normalization and the LeakyReLU activation function.
We pad the features with size $\lfloor\nicefrac{k}{2}\rfloor$ to retain the spatial dimensions of the input.

\myparagraph{Message passing}
The output of curve and surface CNNs are hidden features treated as input edge and node features to the graph neural network.
Given the initial features, we compute the hidden node features $h_v^{(k)}$ in graph layer $k\in 1\ldots K$, by aggregating the input node features $h_v^{(k-1)}$ from a one-hop neighborhood $u \in N(v)$ while conditioning them on the edge features $h_{uv}^{(k-1)}$:

\begin{align}
h_v^{(k)} = \phi^{(k)}\bigg(&(1 + \epsilon^{(k)})~h_v^{(k-1)} +\nonumber\\
                            &\sum_{u \in N(v)} f_\Theta \big(h_{uv}^{(k-1)}\big) \odot h_u^{(k-1)} \big)
                      \bigg),
\label{eq:node_message_passing}
\end{align}
where $\phi^{(k)}$ is a multi-layer perceptron (MLP) with two fully connected layers $FC(64, 64)\rightarrow FC(64, 64)$, $\epsilon^{(k)}$ is a learnable parameter to distinguish the center nodes from the neighbors and $f_\Theta$ is a linear projection from the edge to node feature space.
This update equation extends the Graph Isomorphism Network~\cite{xu2018:gin}, with additional consideration of edge features.
The hidden edge features are next updated similarly while considering the features of the endpoint nodes:
\begin{equation}
h_{uv}^{(k)} = \psi^{(k)}\bigg( (1 + \gamma^{(k)})~h_{uv}^{(k-1)} + f_\Xi\big(h_{u}^{(k-1)} + h_{v}^{(k-1)}\big) \bigg),
\label{eq:edge_message_passing}
\end{equation}
where $\psi^{(k)}$ is a 2-layer MLP as before, $\gamma^{(k)}$ is a learnable parameter to distinguish the edge features from its neighbors, and $f_\Xi$ is a linear projection from the node to edge feature space.
At the end, we then take all the hidden node features $\{ h_v^{(k)}~|~ k \in 1\ldots K \}$ and apply an element-wise max-pooling operation across the nodes to obtain hierarchical graph-level feature vectors from every layer $\{ h^{(k)}~|~k \in 1\ldots K\}$, where $h^{(k)} = {\textrm{maxpool}_{v \in V}(h_v^{(k)})}$.
These features are then linearly projected into 128D vectors and summed to obtain the final shape embedding:
\begin{equation}
        h_G = \sum_{k=1}^K w^{(k)} \cdot h^{(k)} + b^{(k)} .
\end{equation}

We use $K=2$ graph layers in all experiments.
The node and graph embeddings obtained from the network can be used for several downstream applications as detailed next.
\section{Experiments}
\label{sec:experiments}
In this section, we qualitatively and quantitatively evaluate UV-Net on 3D shape classification, segmentation, and shape retrieval on unlabelled data.

\subsection{Datasets}
We briefly introduce the five datasets used in our experiments and provide further details in the supplementary material. We select the datasets below as they are available in B-rep format, unlike many common benchmark datasets provided in mesh format.

\myparagraph{Machining feature dataset~\cite{zhang2018:featurenet}} a synthetic labeled, balanced dataset representing machining features such as chamfers and circular end pockets applied to a cube. It has 23,995 3D shapes ($\sim$1000 per class) split into 24 classes.

\myparagraph{MFCAD dataset~\cite{cao2020:gnn}} a synthetic segmentation dataset of 15,488 3D shapes, similar to the Machining feature dataset, but with multiple machining features. 16 different segmentation labels are used and applied per face.

\myparagraph{FabWave dataset~\cite{atin2019:fabsearch}} a small labeled, imbalanced collection of 5,373 3D shapes split into 52 mechanical part classes, such as brackets, gears, and o-rings.

\myparagraph{ABC dataset~\cite{koch2019:abc}} a real-world collection of millions of 3D shapes.
The dataset is unlabelled, imbalanced, and has many duplicates.
We remove duplicates and use a subset of 46k models in our experiments.

\myparagraph{SolidLetters dataset} our dataset consists of 96k 3D shapes generated by randomly extruding and filleting the 26 alphabet letters (a--z) to form class categories across 2002 style categories from fonts.
Compared to other synthetic datasets that have similar intra-class topology, SolidLetters contains significant variations in both geometry and topology, due to font variety, and is well-balanced.

\subsection{Tasks}
We now compare UV-Net to PointNet~\cite{qi2017:pointnet}, DGCNN~\cite{wang2019:dgcnn}, and MeshCNN~\cite{hanocka2019:meshcnn} on several standard tasks. We show additional results from the baseline methods presented with the Machining feature and MFCAD datasets.

\subsubsection{Classification}
We first evaluate our method on the task of 3D shape classification. The ability to classify 3D components in large B-rep assemblies is valuable for numerous applications including product lifecycle management and automation of repetitive tasks such as simulation setup. 
We show the advantages of using both geometry and topology in the B-rep.
This is particularly important in datasets where data within a class has high geometric variance but similar topology, as is common in parametric CAD modeling.
Our network comprises the UV-Net encoder network in Figure~\ref{fig:network_arch} followed by a non-linear classifier (2-layer MLP) that maps the 128D shape embedding into class logits.
Our input geometric features include xyz coordinates and the trimming mask.

We train point cloud-based methods on 2048 points sampled  uniformly from the solid model, 
FeatureNet~\cite{zhang2018:featurenet}, the baseline for the Machining feature dataset, on 64$^3$ voxel grids, and MeshCNN~\cite{hanocka2019:meshcnn} on triangle meshes.
For the Machining Feature dataset we convert B-reps into high-quality, watertight, manifold meshes as required by MeshCNN using the finite-element mesher in Autodesk Fusion 360 with a target edge-count of 2000 edges.  As MeshCNN requires all meshes to have a similar edge count, we find it is impractical to use with datasets of varying shape complexity, such as FabWave, SolidLetters, and ABC. Although it may be feasible to use a target edge count suitable for the most complex shape in the dataset, in practice this dramatically increases training time and limits the advantages of mesh pooling.

We train all models to a maximum of 350 epochs with cross-entropy loss and the Adam~\cite{kingma2014:adam} optimizer.
Table~\ref{tab:classification} shows that our method achieves the best classification accuracy on all datasets.
Unstructured representations suffer when data within a class has high geometric variance but similar topology, since they cannot model the latter explicitly.
Notably, we outperform FeatureNet~\cite{zhang2018:featurenet} on their dataset, and obtain the highest results on SolidLetters, demonstrating that our method can exploit both geometry and topology.
\begin{table}
\small
\centering
\caption{Solid model classification.}
\begin{tabular}{llcr}
\toprule
Dataset                         &Model                    &Accuracy (\%)                &\#Param. \\
\midrule
\multirow{4}{1.5cm}{Machining Feature}  &UV-Net               & {\bf 99.94 $\pm$ 0.00}  & 1.34M \\
                                    &PointNet (2048)      & 87.13 $\pm$ 0.15            & 0.81M\\
                                    &DGCNN (2048)         & 92.81 $\pm$ 0.69            & 1.81M\\
                                    &FeatureNet (64$^3$)  & 98.85 $\pm$ 0.48            & 33.94M\\
                                    &MeshCNN (2000)       & 98.90 $\pm$ 0.70            & 0.67M\\                                    
\midrule
\multirow{3}{1.5cm}{FabWave}        &UV-Net               & {\bf 94.51 $\pm$ 0.10}      & 1.35M       \\
                                    &PointNet (2048)      & 80.08 $\pm$ 3.61            & 0.82M\\
                                    &DGCNN (2048)         & 69.95 $\pm$ 2.37            & 1.81M\\
\midrule
\multirow{3}{1.5cm}{SolidLetters}   &UV-Net               & {\bf 97.24 $\pm$ 0.10}      & 1.34M\\
                                    &PointNet (2048)      & 94.72 $\pm$ 0.17            & 0.81M\\
                                    &DGCNN (2048)         & 96.62 $\pm$ 0.13            & 1.81M\\
\bottomrule
\end{tabular}
\label{tab:classification}
\end{table}


\subsubsection{Segmentation}
We now consider the problem of segmenting the faces of a B-rep, a classic task with applications in machining feature recognition,  computer-aided process planning and CAD modeling history reconstruction.
We consider the MFCAD and ABC datasets in this experiment and demonstrate the benefit of directly working with B-rep entities.
To work around lack of labels in the ABC dataset, we use the Autodesk Shape Manager~\cite{asm}, a commercial solid-modeling kernel, to assign labels indicating the CAD operation likely to have created the face, such as \textit{ExtrudeSide}, \textit{ExtrudeEnd}, or \textit{Fillet}, we provide more details in supplementary material.

Our segmentation network is similar to the classification network with a small difference: we concatenate the shape embedding to each of the node embeddings, and use a non-linear classifier to output per-node logits.
We additionally include curve tangents and surface normals in the edge and face input features, respectively.

To investigate the benefits of working with B-rep data directly, we compare against established point and mesh-based methods.
We mesh the B-reps in the MFCAD dataset, as previously described and discard $27$ shapes that fail to mesh.
To generate point clouds for both the MFCAD and ABC datasets, we first convert the B-reps into render-meshes, i.e., non-watertight, disjoint meshes.
We then sample the triangles uniformly based on the surface area to generate 2048 points.
The mapping between the faces and primitives (edges, and points) are retained, so that we can perform a per-face voting to compute face-level scores.

\begin{table*}
    \centering
    \caption{Solid face segmentation. The per-primitive scores refer to per-point scores for point cloud-based models and per-edge scores for MeshCNN. The corresponding per-face scores are computed by voting the predictions from all primitives in a face.}
    \small
    \begin{tabular}{llccccccc}
    \toprule
    \multirow{2}{*}{Dataset}    & \multirow{2}{*}{Model}   &\multicolumn{2}{c}{Accuracy}   &\multicolumn{2}{c}{Per-class accuracy}   &\multicolumn{2}{c}{Intersection-over-Union}  & \multirow{2}{*}{\#Param.}\\
    \cmidrule{3-8}
                             &                       &Per-face  &Per-prim.                     &Per-face   &Per-prim.         &Per-face    &Per-prim.\\
    \midrule
    \multirow{6}{*}{MFCAD}   & UV-Net                & {\bf 99.95 $\pm$ 0.02}&-               & {\bf 99.93 $\pm$ 0.20}&-            & {\bf 99.87 $\pm$ 0.03}&- &1.23M\\
                             & UV-Net (xyz)          & 99.83 $\pm$ 0.06&-               & 99.80 $\pm$ 0.00 &-   & 99.63 $\pm$ 0.06 &- &1.23M\\
                             & PointNet              &32.13 $\pm$ 7.92 &59.13 $\pm$ 7.54                                        &16.20 $\pm$ 8.51 & 15.78 $\pm$ 8.17                             &7.15 $\pm$ 5.22 &8.27 $\pm$ 4.66 &0.87M\\
                             & DGCNN                 &82.50 $\pm$ 2.46 &91.60 $\pm$ 2.18   &80.43 $\pm$ 4.51 &78.80 $\pm$ 4.57                                     &67.70 $\pm$ 4.73 &78.67 $\pm$ 6.27 &0.98M\\
                             & GNN                   &-&-                            &-&-                           &93.60~\cite{cao2020:gnn}&- &0.53M\\
                             & MeshCNN               & 99.89 $\pm$ 0.01  & 98.52 $\pm$ 0.04    & 99.84 $\pm$ 0.03 & 98.29 $\pm$ 0.09                  & 99.70 $\pm$ 0.06 & 95.93 $\pm$ 0.05 &2.29M\\
    \midrule
    \multirow{4}{*}{ABC}     & UV-Net &{\bf 88.87 $\pm$ 0.70} &- &{\bf 56.81 $\pm$ 0.93} &- &{\bf 50.37 $\pm$ 1.11} &- &1.23M\\
                             & UV-Net (xyz) &77.33 $\pm$ 0.48 &- &47.38 $\pm$ 0.54 &- & 38.99 $\pm$ 0.42 &- &1.23M\\
                             & PointNet &40.77 $\pm$ 1.79 &61.27 $\pm$ 0.55 &19.87 $\pm$ 0.51 &25.53 $\pm$ 0.32 &11.10 $\pm$ 0.70 &18.47 $\pm$ 0.31 & 0.87M\\
                             & DGCNN &54.18 $\pm$ 3.19 &67.80 $\pm$ 0.59 &27.30 $\pm$ 1.34 &34.93 $\pm$ 1.52 &18.14 $\pm$ 1.97 &26.26 $\pm$ 1.27 &0.98M\\
    \bottomrule
    \end{tabular}
    \label{tab:segmentation}
\end{table*}
We train the models as before using the cross-entropy loss.
Considering each face in the B-rep as a data point, we report the accuracy, per-class accuracy and intersection-over-union (IoU) metrics in Table~\ref{tab:segmentation}.
Results show that UV-Net solves the face segmentation problem in the MFCAD dataset, outperforming their baseline method~\cite{cao2020:gnn} and point cloud-based methods by a wide margin.
MeshCNN~\cite{hanocka2019:meshcnn} obtains very similar results to UV-Net; we suggest this is due to the dihedral angle feature used, which many segmentation labels in the MFCAD dataset strongly correlate with.
Our method achieves state-of-the-art results while operating on B-reps directly and avoids the problem of producing consistent meshing.
We observe a similar trend with the ABC dataset.
Point cloud methods are unable to discover the topological information necessary to identify rare classes, and suffer from loss of fidelity.

\subsubsection{Self-supervised learning}
\label{ssec:clr}
Learning from unlabeled data is important with solid models since real-world labeled datasets are limited, and representation learning by an encoder-decoder scheme is non-trivial due to a lack of B-rep decoders.
We leverage contrastive learning~\cite{wu2018:unsupervised,chen2020:simclr2,he2020:moco,sanghi2020:info3d} (CLR) and propose the following transformations to create positive views for training on B-rep data, each of which enforces a useful prior on the model.

\myparagraph{Connected patch} Extract a random node and its $n$-hop neighbors ($n\in\{1, 2\}$). This implies that local patches in a B-rep hint about the global shape.

\myparagraph{Drop nodes} Randomly delete nodes with uniform probability (0.4) along with attached edges to encourage B-reps with partially similar faces to be clustered together in the latent space.

\myparagraph{Drop edges} Randomly delete edges with uniform probability (0.4), to encourage B-reps that look similar visually, but have different topology be clustered together.

\begin{figure}  
    \centering
    \includegraphics[width=0.95\columnwidth]{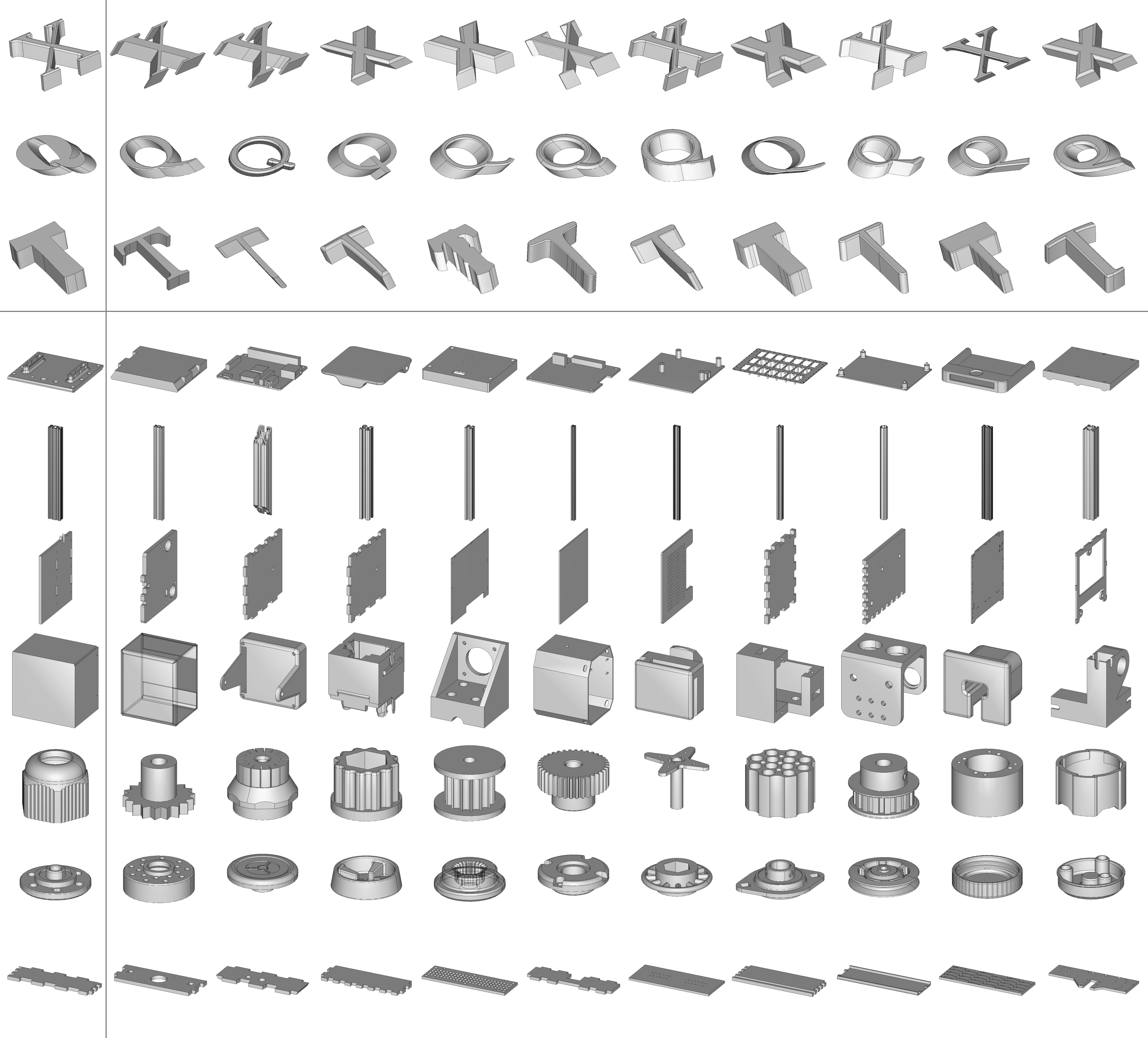}
    \caption{Self-supervised shape retrieval on SolidLetters and ABC datasets. Column 1: Query, Columns 2--11: Retrieved results sorted left to right by distance in latent space.}
    \label{fig:retrieval}
\end{figure}

Our CLR model has three components~\cite{chen2020:simclr2}.
Given a B-rep in UV-grid+graph representation $G$, we uniformly sample two i.i.d. transformations $T_1$ and $T_2$ to obtain two positive views $T_1(G)$ and $T_2(G)$.
Occasionally (10\% of the time), we set $T_1$ to the identity transformation so that the global shape is available to the neural network to associate with the other partial views.
An ablation study for these transformations is provided in the supplementary material.

Our UV-Net encoder extracts the 128D shape embeddings $h_i$ and $h_j$ of positive pairs.
A 3-layer non-linear projection head (MLP) with ReLU activations maps these embeddings to 64D latent vectors $z_i$ and $z_j$.
Given a mini-batch of size $N=256$, we compute $\{z_k~|~k\in 1 \ldots 2N\}$, and bring together the positive pairs while treating the remaining $2(N - 1)$ data as negative examples, with the normalized temperature scaled
cross-entropy~\cite{chen2020:simclr2} loss.

We first use the SolidLetters dataset (upper case only since CLR performs per-instance discrimination) to quantitatively understand how our method performs.
After training the model for $350$ epochs, we extract the shape embeddings of the test set and perform k-means clustering to generate $26$ clusters.
We measure the clustering quality against ground truth clusters (labels) using the adjusted mutual-information metric~\cite{vinh2010:ami}.
We also classify the shape embeddings using a linear Support-Vector Machine (SVM) and compute the classification accuracy.
Results in Table~\ref{tab:clr_solidletters} show that the shape embeddings obtained with our CLR model is rich with category information even though it is trained without labels.
\begin{table}
    \centering
    \caption{Quality of self-supervised shape embeddings obtained with our contrastive learning method on SolidLetters.}
    \small
    \begin{tabular}{lc}
    \toprule
         Method               & Score (\%) \\
    \midrule
         Linear SVM           & 79.40 $\pm$ 0.20\\
         K-means clustering   & 58.17 $\pm$ 0.25\\
    \bottomrule
    \end{tabular}
    \label{tab:clr_solidletters}
\end{table}

To perform shape retrieval, we take random shape embeddings from the test set of SolidLetters and ABC (random 20\% split) as queries, and compute their k-nearest neighbors in the UV-Net shape embedding space as shown in Figure~\ref{fig:retrieval}.
The results demonstrate that our CLR approach is viable, and shows high potential to learn from large-scale unlabeled CAD datasets, an unaddressed problem until now.

\subsection{Sensitivity to sampling}
\label{ssec:sensitivity_sampling}
We now study the effect of the sampling resolution on the accuracy produced by the network.
A robust method should degrade gracefully when the sample count is reduced, or leverage other information to produce consistent results.
\begin{figure}
    \centering
    \includegraphics[width=0.85\columnwidth]{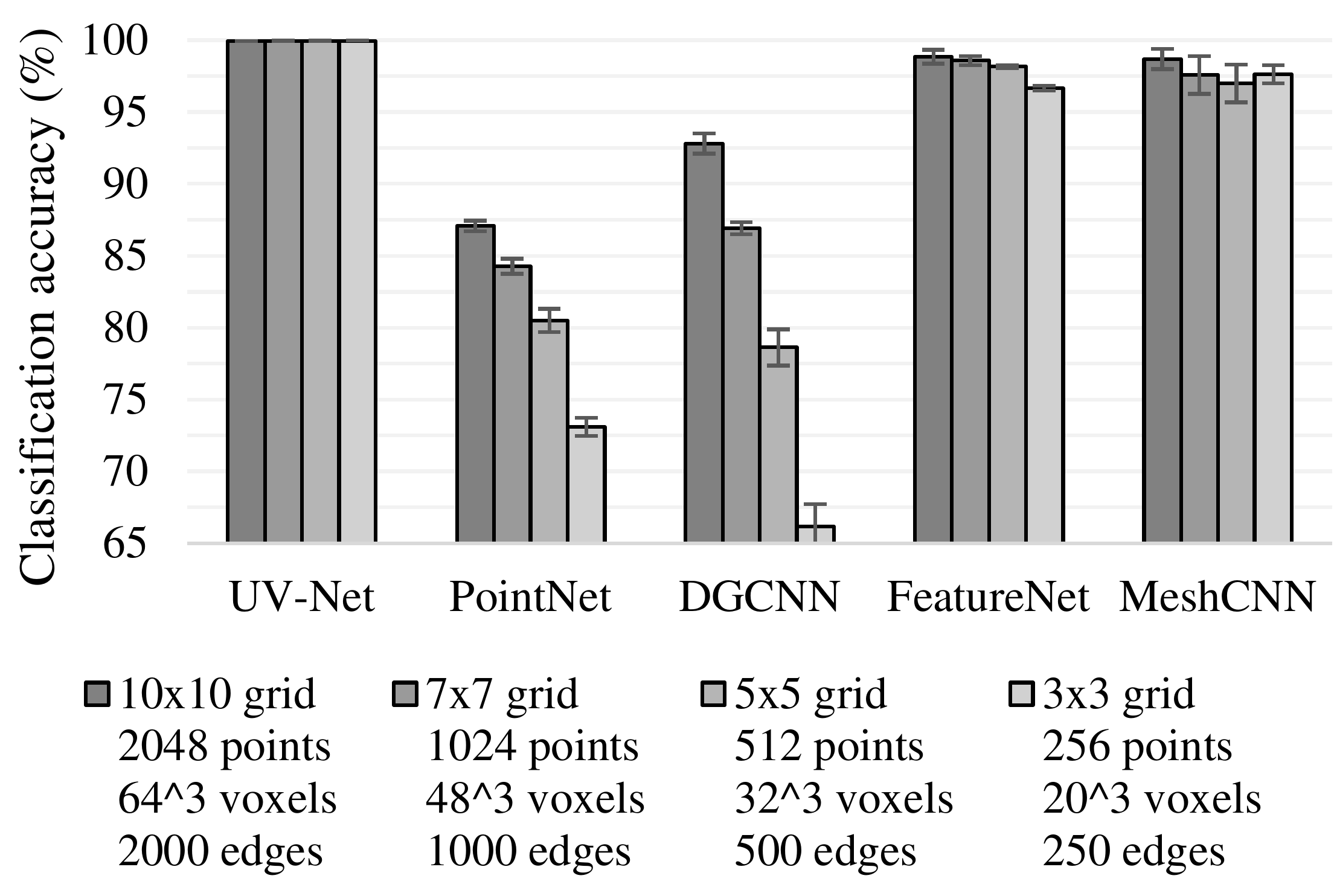}
    \caption{Sensitivity of input representations and methods to sampling resolution for machining feature classification.}
    \label{fig:sensitivity_sampling}
\end{figure}
The classification networks are all trained with reduced resolution data and the accuracy is reported in Figure~\ref{fig:sensitivity_sampling}. 
We see that our method is robust to sampling resolution in the machining feature detection task.
This is because we not only capture the geometry, but also the topological information that can be leveraged for the task.
Moreover, every face in the solid is \textit{seen} by UV-Net regardless of its surface area, while other representations suffer from loss of fidelity.


\subsection{Feature and architecture ablation}
\label{sec:ablation}
Here we study the importance of the input features, and network components on the MFCAD segmentation problem:

\myparagraph{Full UV-Net (xyz only)} We remove the normals from the set of input features but use the full architecture.

\myparagraph{UV-Net (Face only)} This is similar to the full architecture, but without the input curve features and curve CNN ($f_\Theta(h_{uv})$ term in Eq.~\ref{eq:node_message_passing} and entire Eq.~\ref{eq:edge_message_passing} are removed).

\myparagraph{Face features only} We replace the GNN portion of the network with an MLP (similar parameter count as the GNN). Edge features are also removed since they cannot be considered without a message-passing scheme.

\myparagraph{Topology only} We remove the curve and surface CNNs, and set the edge and node attributes of the graph as noise sampled from a normal distribution, so that the network is forced to solve the task with topology features only.

\begin{figure}
    \centering
    \includegraphics[width=0.95\columnwidth]{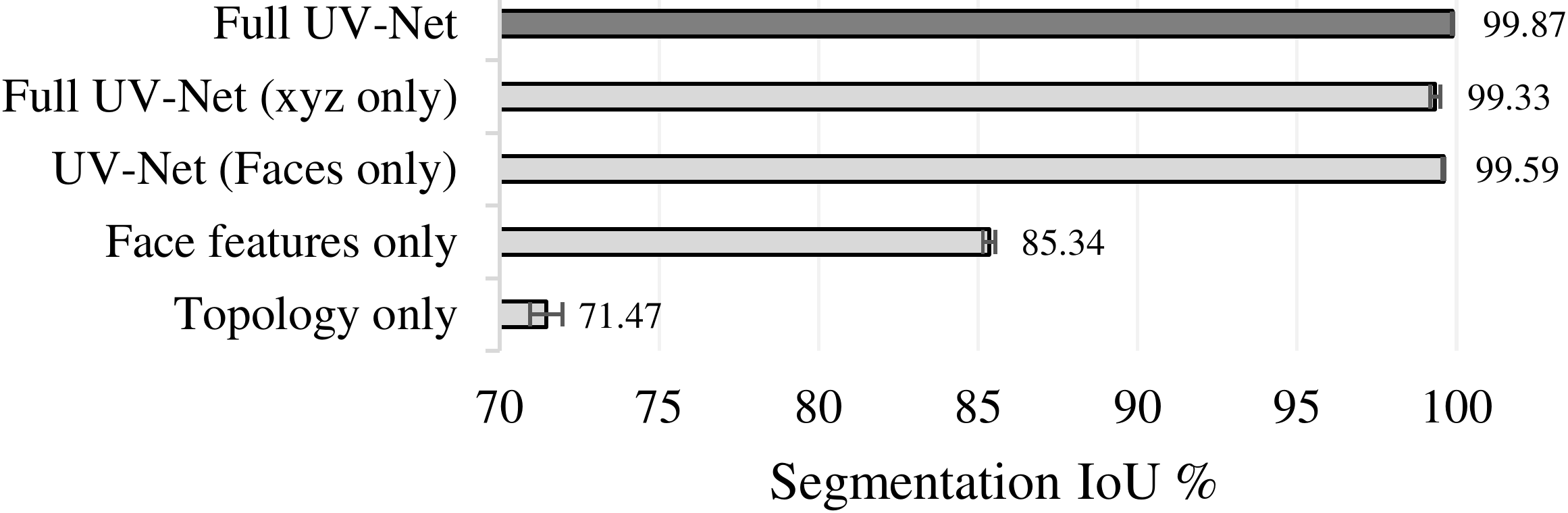}
    \caption{Ablation study with input features and components of UV-Net on the MFCAD segmentation problem.}
    \label{fig:ablation}
\end{figure}
These networks are trained for 100 epochs on the MFCAD segmentation task and the IoU score is compared against the full model.
The benefits of jointly considering the geometric features and topology as proposed is evident from Figure~\ref{fig:ablation}, and validates the merits of our approach.


\subsection{Invariance to reparametrization}
A solid can be altered in subtle ways without changing the 3D appearance by reparametrizing the curve/surface geometry.
This can occur when converting models from one format to another (e.g. STEP to SAT), changing the surface type (e.g. plane to spline), or as a result of some high level CAD modeling operation.
The UV-grid is to a large extent invariant to common reparametrizations as discussed earlier.

On the other hand, reversing a surface parametrization along the u- or v-axis amounts to flipping the UV-grid about the same axis, while transposing the surface parametrization by exchanging the u- and v-axis is equivalent to rotating and flipping the UV-grid.
\begin{table}
    \centering
    \caption{Effect of reparametrizing the SolidLetters classification \textit{test set} on UV-Net.}
    \small
    \label{tab:additional:augmentation}
    \begin{tabular}{lccc}
    \toprule
    Convolution                                    & Reparametrized     &Test Accuracy\\
    \midrule
    \multirow{2}{*}{Regular}                       & No                 &96.74 $\pm$ 0.06\\
                                                   & Yes                &55.98 $\pm$ 2.36\\
    \midrule
    \multirow{2}{*}{$\text{D}_2$ equivariant}      & No                 &96.58 $\pm$ 0.01\\
                                                   & Yes                &96.59 $\pm$ 0.02\\
    \bottomrule
    \end{tabular}
\label{tab:augmentation}
\end{table}
Flips about u- and v-axis and rotations by $\{k\frac{\pi}{2}~|~k \in [0, 1, 2, 3]\}$ belong to the Dihedral symmetry group $\text{D}_2$, and regular image convolutions are not invariant to them as we show in Table~\ref{tab:augmentation}.
We see that randomly performing these transformations to surfaces in the test set but not the training set affects the classification accuracy.
However, employing $\text{D}_2$ group equivariant convolutions~\cite{weiler2019:e2cnn} followed by a group pooling layer in the surface CNN~(Section~\ref{sec:input_representation}) makes the model resilient to these reparametrizations. 

\section{Conclusion}
We have presented UV-Net, a neural network and representation that can work on B-rep data, and leverage existing image and graph convolutional neural networks.
We have shown its benefits and versatility on both supervised and self-supervised tasks spanning five B-rep datasets, outperforming other representations such as point clouds, voxels, and meshes.
In addition, we introduced SolidLetters, a new synthetic B-rep dataset with variations in both geometry and topology.
We believe our work can unlock data-driven applications in established CAD modeling pipelines, and revitalize research interest in this domain.

\myparagraph{Limitations \& future work}
We fixed the sampling step size for each curve or surface regardless of its geometry.
Choosing the step using derivatives~\cite{zheng2000:estimatetesselation,kawasaki2018:imageprocbspline} or learning it in a task-dependent manner could be an interesting extension.
While UV-grids are versatile, we did not exploit other information available in the B-rep such as curve and surface types, edge convexity, halfedge ordering, etc. which might prove useful in certain applications.
Finally, our UV-grid features are not rotation-invariant. Although we can use local coordinates for each UV-grid~\cite{deng2018:ppfnet,deng2018:ppffoldnet} or switch to other features like mean-curvature, this may make the network lose sight of the relative orientation of various faces and edges. We leave the detailed study of various invariances to future work.
We also believe there is tremendous potential to improve our self-supervised method for transfer learning from large datasets like ABC.
Finally, it is worth investigating how ideas from this work can be adapted to other representations like subdivision surfaces, where the limit surface can be parametrized as a regular structure using the faces of the control mesh.

\appendix
\counterwithin{figure}{section}
\counterwithin{table}{section}

\section{Supplementary material}

\subsection{Sampling error in UV-grids}
\label{app:sampling}
We quantitatively measure how well the UV-grids approximate the original surface geometry. We chose a random selection of 132,492 models from the ABC dataset, and converted them into our UV-grid representation: curves are represented by grids with 10 points and unit tangents, while surface are represented by 10$\times$10 grids with points and unit surface normals.
We then compute four metrics, two related to edge curve approximation and two related to surface approximation:
\begin{itemize}
\item
Chordal error (curves): The distance between the center of the line joining two points and the ground truth curve evaluated at the average $u$ parameter value of two successive sample points.
\item
Chordal error (surfaces): The distance between the average of four points defining a patch on a 10$\times$10 point grid and the real surface evaluated at the average $(u,v)$ of the patch. This error metric is considering the point grid as a bi-linear approximation of the surface.
\item
\bez approximation error (curves): A cubic \bez{} span is constructed from the points and unit tangent vectors following Equation 9.47 in \cite{piegl1996:nurbsbook}.  The \bez approximation error is then taken as the average distance between the center of the \bez{} and the real edge curve evaluated at the average $u$ parameter value of the two sample points used to construct the \bez span.
\item
\bez approximation error (surfaces): A cubic \bez{} patch is constructed from the 4 points and unit normals following Equation 9.58 in \cite{piegl1996:nurbsbook}. The \bez approximation error is then taken as the average  distance between the center of the patch, and the real surface evaluated the central $(u,v)$ parameter value. This error metric is considering the point grid as a cubic \bez approximation of the surface.
\end{itemize}
To allow these errors to be compared for solids of different sizes we divide each by then longest length of the bounding box of the entire solid. 

\begin{table}[b]
    \centering
    \caption{Percentage of curves and surfaces with approximation errors exceeding various thresholds computed on random samples from the ABC dataset.}
    \small
    \begin{tabular}{lcccc}
    \toprule
    \multirow{2}{1.5cm}{Factor of box size}    & \multicolumn{2}{c}{Surfaces} & \multicolumn{2}{c}{Curves} \\
    \cmidrule{2-5}
    & \bez & Chordal & \bez & Chordal\\
    \midrule
    Above $10^{-3}$ & 3.16\% & 10.81\% & 0.23\% & 6.67\% \\
    Above $10^{-2}$ & 0.80\% & 2.65\% & 0.06\% & 1.33\% \\
    Above $10^{-1}$ & 0.09\%& 0.10\%& 0.02\%& 0.02\%\\
    \bottomrule
    \end{tabular}
    \label{tab:uvgrid-sampling-error}
\end{table}
As the neural network is passed ordered lists of edge curve points and tangents and an ordered grid of points and normals, the network has sufficient information to understand the curve and surface information as a linear/bilinear interpolation. Chordal errors of \textbf{89.19\%} surface patches and \textbf{93.33\%} curves are within $10^{-3}$ of the longest length of the B-rep's bounding box as shown in Table~\ref{tab:uvgrid-sampling-error}.

The network also has access to curve tangent and surface normal information. If we assume that the surface normal information can be used by the network then the approximation error is further reduced and we find that the \bez approximation errors of \textbf{96.84\%} surface patches and \textbf{99.77\%} are within $10^{-3}$ of the longest length of the B-rep's bounding box.

While it is unclear if the network actually uses this information to build an interpolation of the curve/surface geometry, this information is part of the input and is empirically found to help in our ablation studies (see Section~\ref{sec:ablation}).

\subsection{SolidLetters dataset}
\label{app:dataset}
A publicly available, balanced, and labeled dataset is vital to assist in designing and testing B-rep neural network architectures.
To this end, we create ``SolidLetters'', a new, synthetic, labeled dataset for solid models that includes both geometric and topological variations.
It comprises upper and lower case letters in various styles obtained from a collection of 2002 system and Google Fonts.
Each data point has three labels: (1) the alphabet, (2) the case (upper or lower), and (3) the name of the font.
\begin{figure}[b]
    \centering
    \begin{subfigure}[b]{0.22\textwidth}
        \centering
        \includegraphics[height=0.49\textwidth, keepaspectratio]{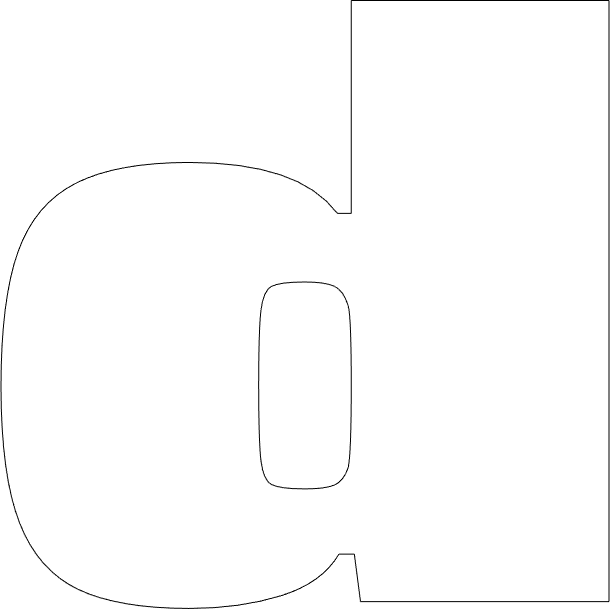}
        \caption{}\label{fig:solid_mnist_running_example_a}
    \end{subfigure}
    \begin{subfigure}[b]{0.24\textwidth}
        \centering
        \includegraphics[height=0.4\textwidth, keepaspectratio]{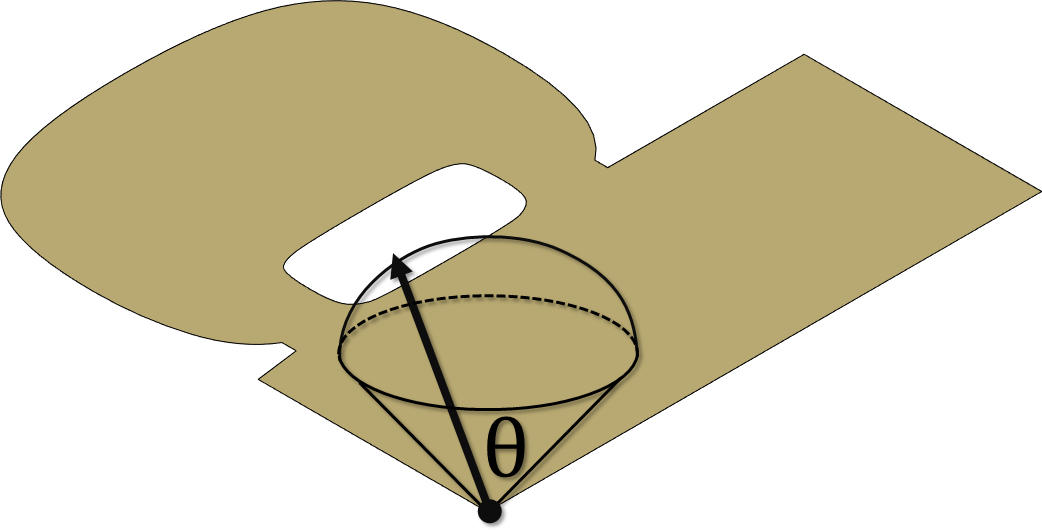}
        \caption{}\label{fig:solid_mnist_running_example_b}
    \end{subfigure}
    \\
    \begin{subfigure}[b]{0.23\textwidth}
        \centering
        \includegraphics[height=0.49\textwidth, keepaspectratio]{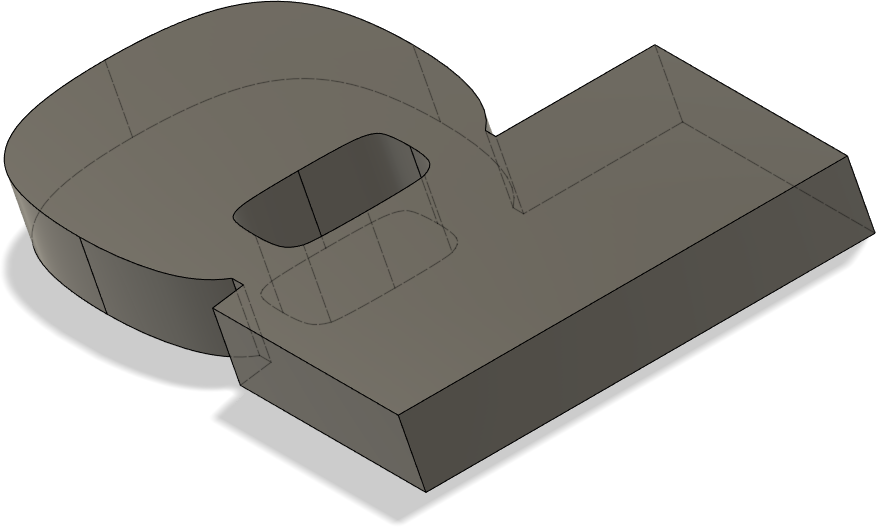}
        \caption{}\label{fig:solid_mnist_running_example_c}
    \end{subfigure}
    \begin{subfigure}[b]{0.23\textwidth}
        \centering
        \includegraphics[height=0.49\textwidth, keepaspectratio]{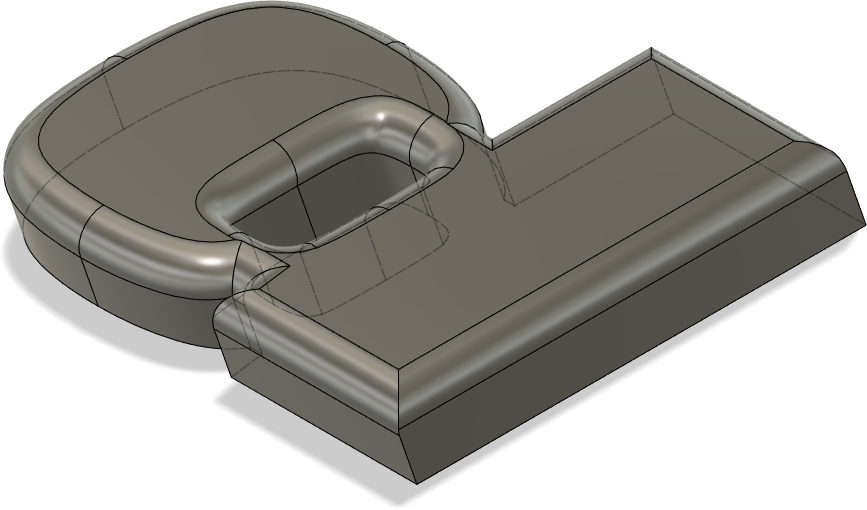}
        \caption{}\label{fig:solid_mnist_running_example_d}
    \end{subfigure}
    \caption{Running example of data generation. (a) 2D wire B-rep going through boundary of the font face. (b) Trimmed planar sheet filling the interior of the boundary. (c) Extrude. (d) Fillet edges of the topmost face (SolidLetters).}
    \label{fig:solid_mnist_running_example}
\end{figure}

\begin{figure*}
    \centering
    \includegraphics[width=0.99\textwidth]{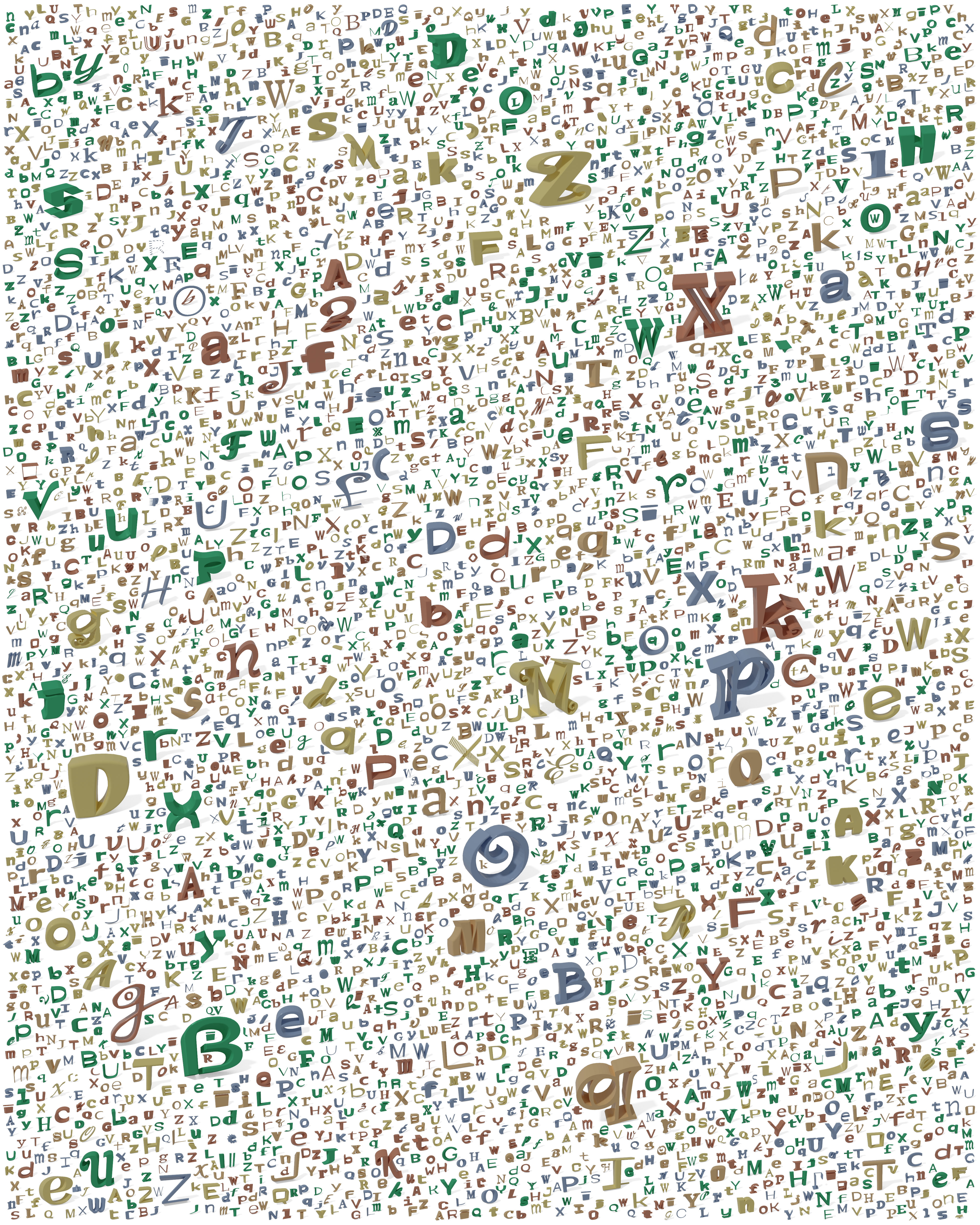}
    \caption{Visual overview of the SolidLetters dataset.}
    \label{fig:SolidLetters_mosaic}
\end{figure*}

\myparagraph{Creation}
We first create the outline of each letter with every font (size 10) (Figure~\ref{fig:solid_mnist_running_example_a}), and fill its interior with a trimmed planar sheet surface, see Figure~\ref{fig:solid_mnist_running_example_b}.
Treating the planar sheet as a profile surface on the XY-plane, we extrude it along a vector $\mathbf{e}$ pointing upwards, see Figure~\ref{fig:solid_mnist_running_example_c}.
We define this vector such that its head lies at a random point in the spherical cap situated along the z-axis to introduce variance in the extrusion direction.
By sampling two random numbers $\xi_1$ and $\xi_2$ from a uniform distribution $U(0, 1)$, we can define the vector $\mathbf{e}$ as: $\mathbf{e}_x = \sqrt{1 - \mathbf{e}_z^2} \cos(2\pi \xi_2)$, $\mathbf{e}_y = \sqrt{1 - \mathbf{e}_z^2} \sin(2\pi \xi_2)$, $\mathbf{e}_z = \xi_1 (1 - \cos\theta) + \cos\theta$, where $\theta$ is the angle subtended by the spherical cap that we set to $45^\circ$.
Furthermore, to break the symmetry of the shape across the XY-plane and introduce more complexity in the model, we identify the topmost face in the extruded solid and perform filleting by blending the edges with a constant radius $0.1$, see Figure~\ref{fig:solid_mnist_running_example_d}.
This introduces new curved faces in the model along the edges of the topmost face, and changes the topology as well.
Filleting is prone to failure when the face has edges that meet at sharp angles or the local thickness is small compared to the filleting radius.
Hence, we attempt to fillet three times by successively reducing the filleting radius by 50\%, and upon failure leave the extruded solid as such.
After removing fonts that are non-English and symbols, we end up with a total of 95,795 data points.
\begin{figure}
    \centering
    \includegraphics[width=\columnwidth]{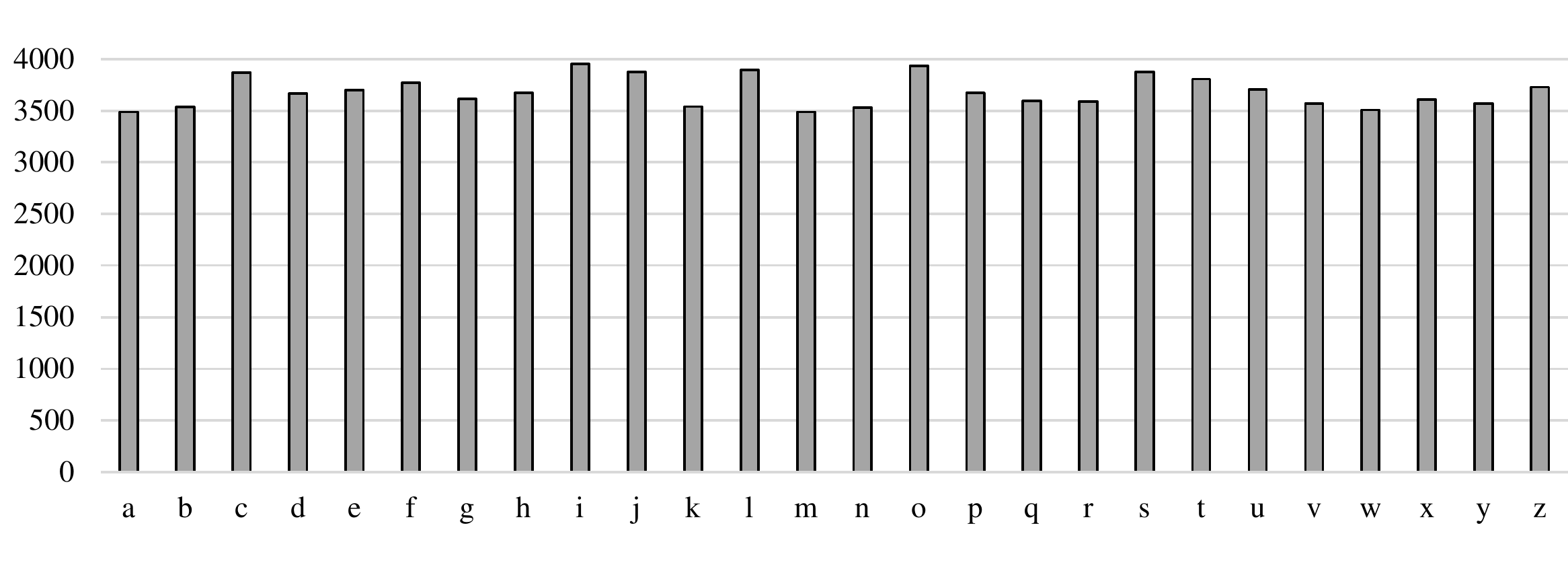}
    \caption{Per-class distribution of solids in the SolidLetters dataset.}
    \label{fig:solidletters_stats}
\end{figure}
There is an average of 33 faces per solid in the dataset.
The per-class distribution of data is shown in Figure~\ref{fig:solidletters_stats}.
We show a visual overview of the entire dataset in Figure~\ref{fig:SolidLetters_mosaic}.

\myparagraph{Data split}
We partition the dataset into an official 80-20 train-test split based on the complexity of the solids, which can be roughly measured using the number of faces.
We place the solids in the datasets into three bins based on the number of faces: $[F_\text{min}, F_1), [F_1, F_2), [F_2, F_\text{max}]$, where $F_\text{min}$ and $F_\text{max}$ are the minimum and maximum number of faces in a solid in the entire dataset, respectively.
$F_1$ is defined as $0.15\times (F_\text{max} - F_\text{min})$, while $F_2$ is set to $0.30\times (F_\text{max} - F_\text{min})$.
The solids in each bin are partitioned randomly into an 80-20 train-test split and finally combined.

\subsection{Other datasets}
\label{app:other-datasets}
\subsubsection{Machining feature}
The Machining feature dataset~\cite{zhang2018:featurenet} is available at \url{github.com/madlabub/Machining-feature-dataset}.
The original train-test split information was not available, so we created a random 85-15 split within each category, and held out 20\% of the training set for validation.

\subsubsection{FabWave}
The FabWave dataset~\cite{atin2019:fabsearch} is available at \url{dimelab.org/fabwave}.
We use the subset of data that the authors call ``Standard" which contains mechanical part categories.
There are a total of 52 part categories, this is 4 less than what is provided in the dataset because we removed some categories that have very few or no models available in them.
There is no official train-test split, so we randomly partitioned the data in a 80-20 ratio within each class.
\begin{figure*}
    \centering
    \includegraphics[width=0.8\linewidth]{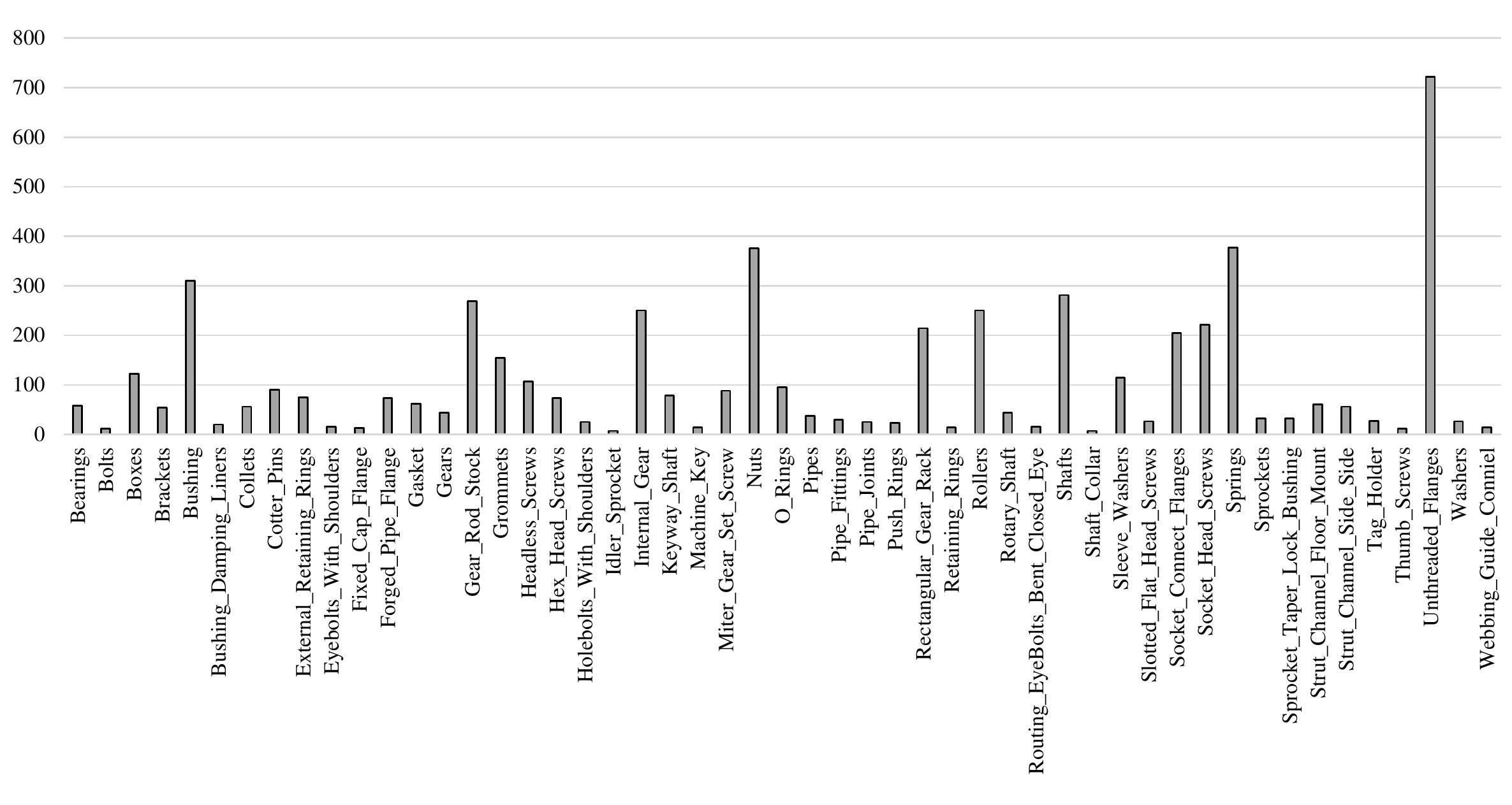}
    \caption{Distribution of categories in the FabWave dataset.}
    \label{fig:fabwave_stats}
\end{figure*}
The data distribution is shown in Figure~\ref{fig:fabwave_stats}.

\subsubsection{MFCAD}
The MFCAD dataset~\cite{cao2020:gnn} is available at \url{github.com/hducg/MFCAD}.
The dataset has 15,488 files (this is 2 more than listed in the paper).
The train-validation-test ratio is 60-20-20, and we use the official split shared by the authors which partitions the models while considering the number of labels per solid.
Unlike the full set of labels described in the paper, the dataset only has labels on planar faces.
This is likely because Cao et al.~\cite{cao2020:gnn}'s method only supports planar faces.
\begin{figure}
    \centering
    \includegraphics[width=\columnwidth]{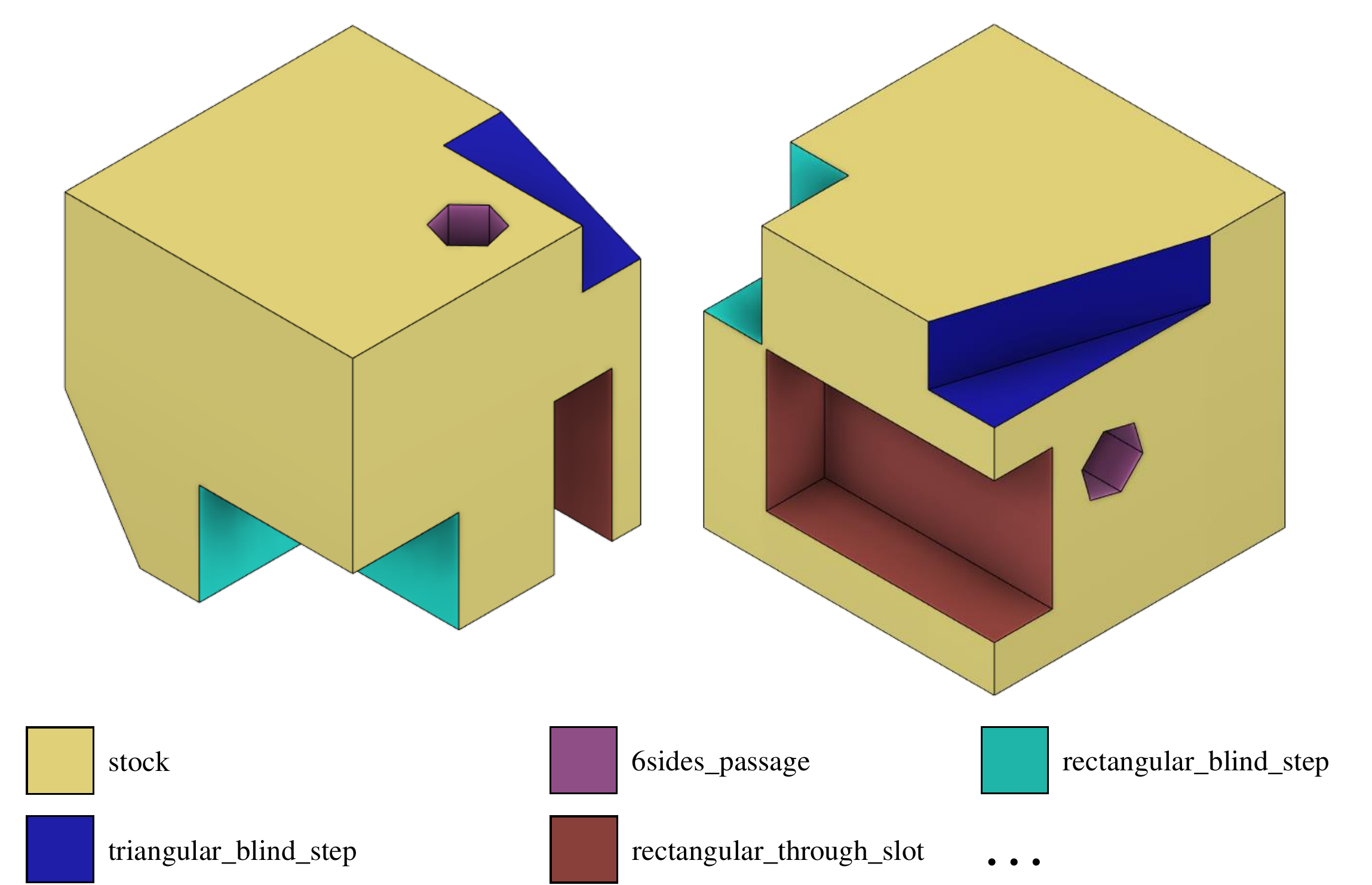}
    \caption{Example 3D models from the MFCAD dataset, colored by segmentation label.}
    \label{fig:mfcadvisual}
\end{figure}
\begin{figure}
    \centering
    \includegraphics[width=\columnwidth]{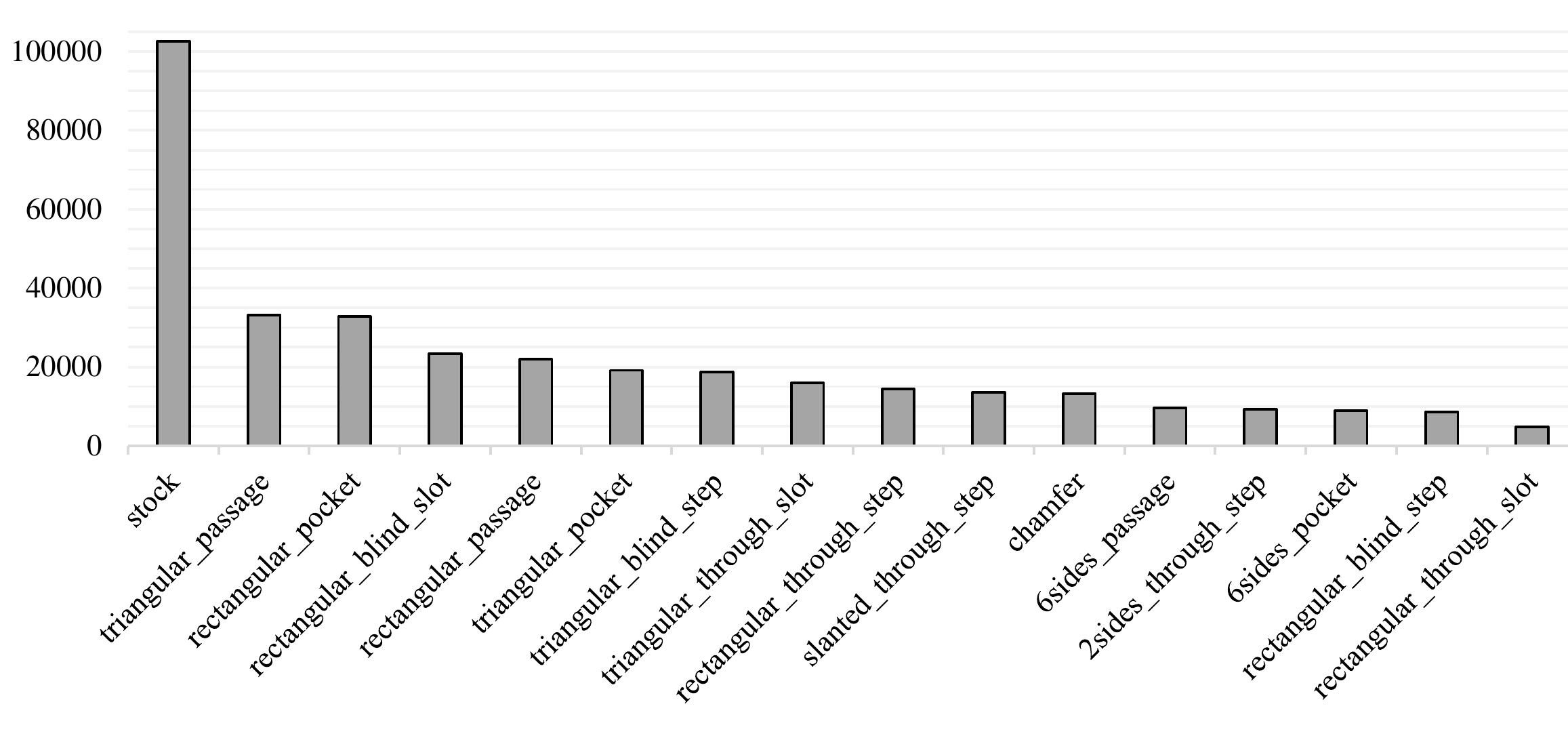}
    \caption{Distribution of segmentation labels in the MFCAD dataset.}
    \label{fig:mfcadstats}
\end{figure}
There are a total of 350,295 faces in the dataset classified into 16 segmentation categories.
Some visual examples are shown in Figure~\ref{fig:mfcadvisual}, and the class distribution in Figure~\ref{fig:mfcadstats}.

\subsubsection{ABC}
The entire ABC dataset~\cite{koch2019:abc} consists of over 1 million CAD assemblies containing over 13 million individual B-rep bodies created by users of the Onshape CAD software. It is available at \url{deep-geometry.github.io/abc-dataset}. To use the dataset in our experiments we use the following process to remove duplicates and generate segmentation labels.

\myparagraph{Duplication removal}
We remove duplicates from the ABC dataset in four steps. All duplicate removal is performed at the B-rep body level, rather than with assemblies.

\begin{enumerate}[leftmargin=*]
    \item \textbf{Remove small files}: A significant number of models in the dataset are simple primitives that are unsuitable for our experiments. We first remove models with a file size of less than 15kB as a simple but effective proxy for removing simple primitives. 
    \item \textbf{Remove file duplicates and invalid files}: We next remove exact file duplicates and invalid file type such as $\mathtt{.xmm\_txt}$.
    \item \textbf{Remove non-solid and simple solids}: We next remove non-solid models, such as those containing only wires or open solids, as well as simple solids with less than 30 faces. 
    \item \textbf{Remove geometric duplicates}: Finally we remove geometric duplicates by creating and comparing a unique hash string for each model using the number of edges, number of faces, number of shells, number of lumps, area, volume, and moments of inertia. This approach is efficient and invariant to rotation. 
\end{enumerate}
From the pool of unique models we choose a random sample of 46k models to use in our experiments.

\myparagraph{Segmentation labels}
Since the ABC dataset is unlabeled, we create our own labels to test UV-Net's segmentation performance on a real-world dataset.
We use the Autodesk Shape Manager (ASM)~\cite{asm} kernel to perform a rule-based feature prediction for each of the faces in the solids.
ASM predicts the modeling operation that could have created the face, e.g., chamfer, fillet, extrude and revolve.
ASM is unable to identify the modeling operation in some cases, and we ignore such faces during training/testing.
Additionally, it also predicts whether the \texttt{change} made by the extrusion was additive or subtractive.
We consolidate this information into labels as follows:
\begin{itemize}[leftmargin=*]
    \item \textit{Chamfer}, \textit{Fillet} and \textit{Revolve} are retained as such. However, we notice that \textit{Chamfer} and \textit{Revolve} are virtually non-existent in our data.
    \item In the case of extrusion, we utilize the extrude direction and the surface normals of a sample of points in the visible region of each face to make a fine-grained categorization.
    \begin{itemize}[leftmargin=*]
        \item If the normals and extrude direction are aligned, then we set the label as \textit{ExtrudeEnd} if the \texttt{change} type is additive, and \textit{CutEnd} if the \texttt{change} type is subtractive.
        \item If the normals and extrude direction are near perpendicular, then we set the label as \textit{ExtrudeSide} if the \texttt{change} type is additive, and \textit{CutSide} if the \texttt{change} type is subtractive.
    \end{itemize}
\end{itemize}
\begin{figure}
    \centering
    \includegraphics[width=\columnwidth]{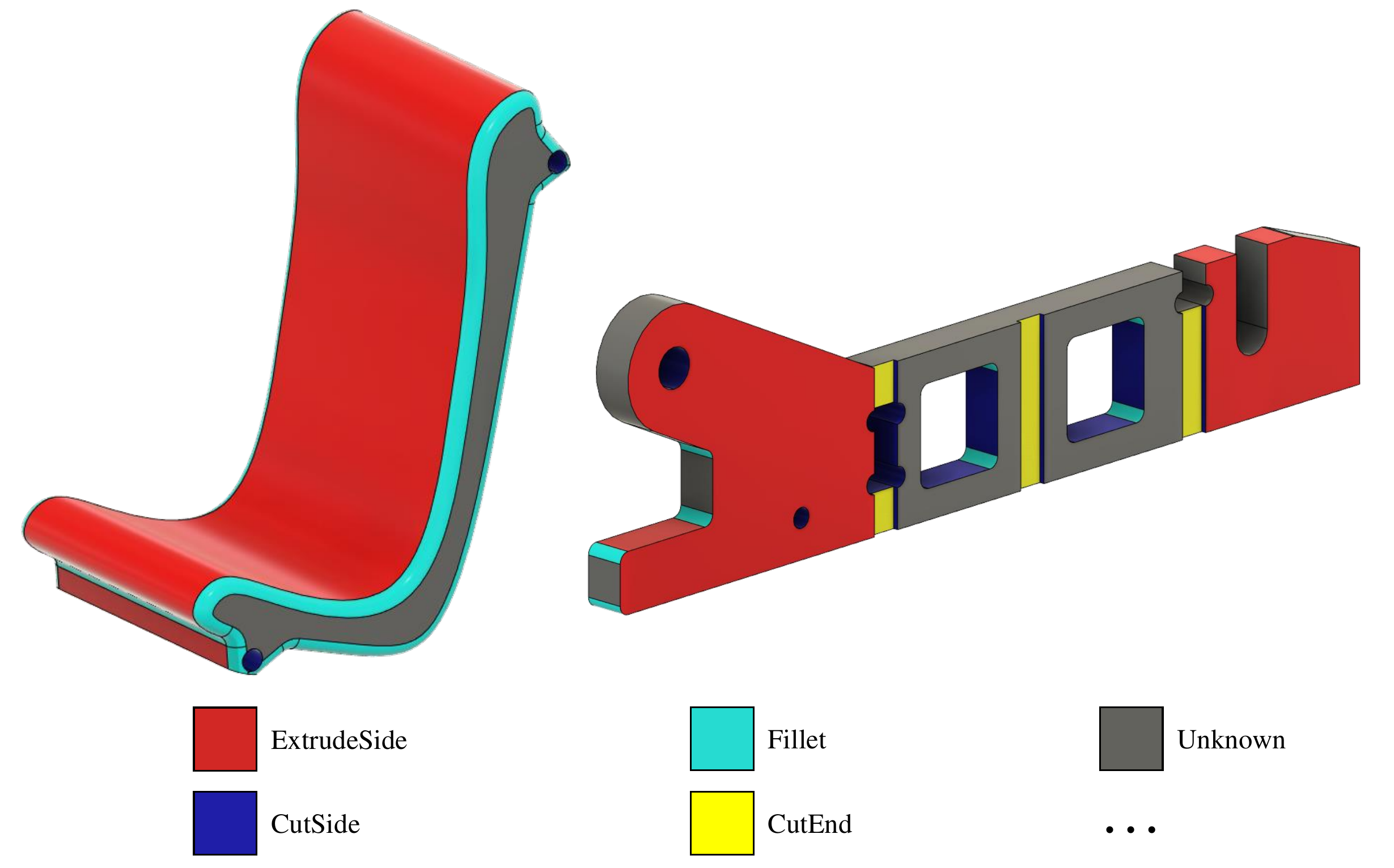}
    \caption{Example 3D models from the ABC dataset, colored by segmentation label.}
    \label{fig:abc_example}
\end{figure}
\begin{figure}
    \centering
    \includegraphics[width=\columnwidth]{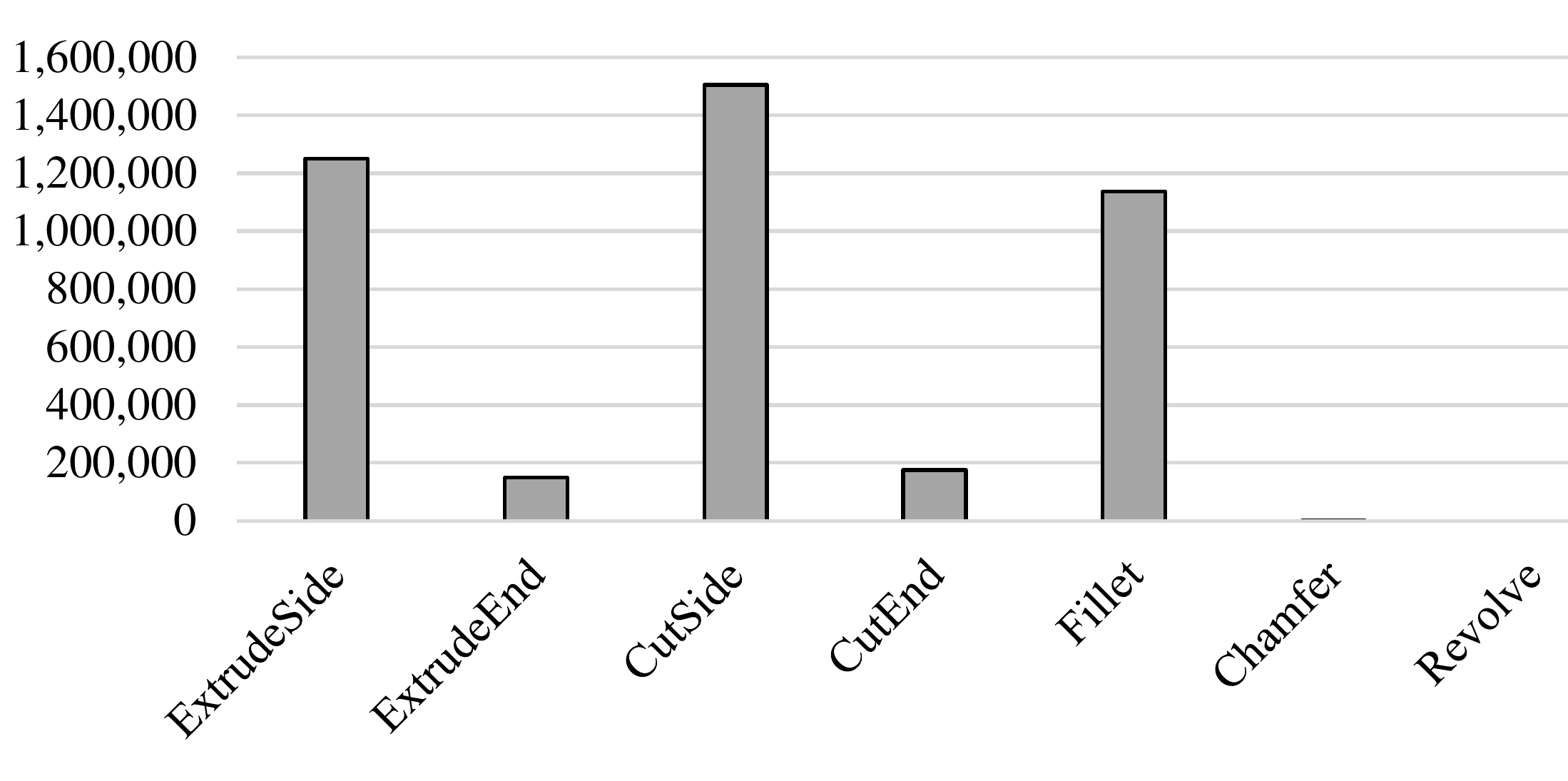}
    \caption{Distribution of segmentation labels in the ABC dataset.}
    \label{fig:abc_stats}
\end{figure}
Our subset of the ABC dataset has a total of 4,218,036 faces.
Figure~\ref{fig:abc_example} shows some example segmentation labels while the label distribution shown in Figure~\ref{fig:abc_stats}.
The dataset is split into train and test sets randomly in a 80-20 ratio.

\subsection{Training details}
Our implementation is in PyTorch and we use DGL (\url{dgl.ai}) for graph operations.
All experiments were conducted on NVIDIA GV100, Quadro P6000, or Tesla V100 GPUs.
All networks in Section~\ref{sec:experiments} are optimized using the Adam optimizer with default parameters (learning rate: 0.001, $\beta_1$: 0.9, $\beta_2$: 0.999).

UV-Net's mini-batches are created by concatenating the nodes and edges of all the graphs in the batch to form a supergraph.
For all classification and segmentation experiments, we used a batch size of 128 for UV-Net, PointNet, and FeatureNet.
We reduced the batch size to 64 for DGCNN, due to its high memory consumption, and used the default mini-batch size of 16 with MeshCNN.
Contrastive learning generally requires a higher batch size since the quality of negative views depend on the data points in the mini-batch, hence, we set it to 256 in this case.
DGCNN has a hyperparameter $k$ to define the number of k-nearest neighbors used to build the graph dynamically in each of its layers.
We set this to 20 in the classification and segmentation experiments.
In the sensitivity to sampling study in Section~\ref{ssec:sensitivity_sampling}, we set $k$ to 10 in the case of $1024$ points, and 5 in the case of $512$ and $256$ points, so that the local neighborhood is well defined.

We adapted the following implementations for our comparisons:
\begin{itemize}[leftmargin=*]
    \item \textbf{PointNet}: we used the PyTorch implementation from the official DGCNN code (\url{github.com/WangYueFt/dgcnn}) for classification, and an unofficial implementation for segmentation (\url{github.com/fxia22/pointnet.pytorch}).
    \item \textbf{DGCNN}: we used the official PyTorch implementation available at \url{github.com/WangYueFt/dgcnn} for classification. Since the segmentation implementation was not available in the official version, we used another implementation from \url{github.com/AnTao97/dgcnn.pytorch} that is recommended by the authors.
    \item \textbf{FeatureNet}: we implemented this model based on the network architecture provided in the paper~\cite{zhang2018:featurenet}.
    \item \textbf{MeshCNN}: we used the official implementation from \url{github.com/ranahanocka/MeshCNN}.
\end{itemize}

\subsection{Additional self-supervised results}
\subsubsection{Ablation on CLR transformations}
We perform an ablation study on the different transformations that we proposed to generate views for contrastive learning.
We train our CLR model on the SolidLetters dataset for 100 epochs while removing one transformation at a time.
While training for 100 epochs is not sufficient for the network to converge, it gives us a fair understanding of the importance of each transformation.
\begin{figure}
    \centering
    \includegraphics[width=0.98\columnwidth]{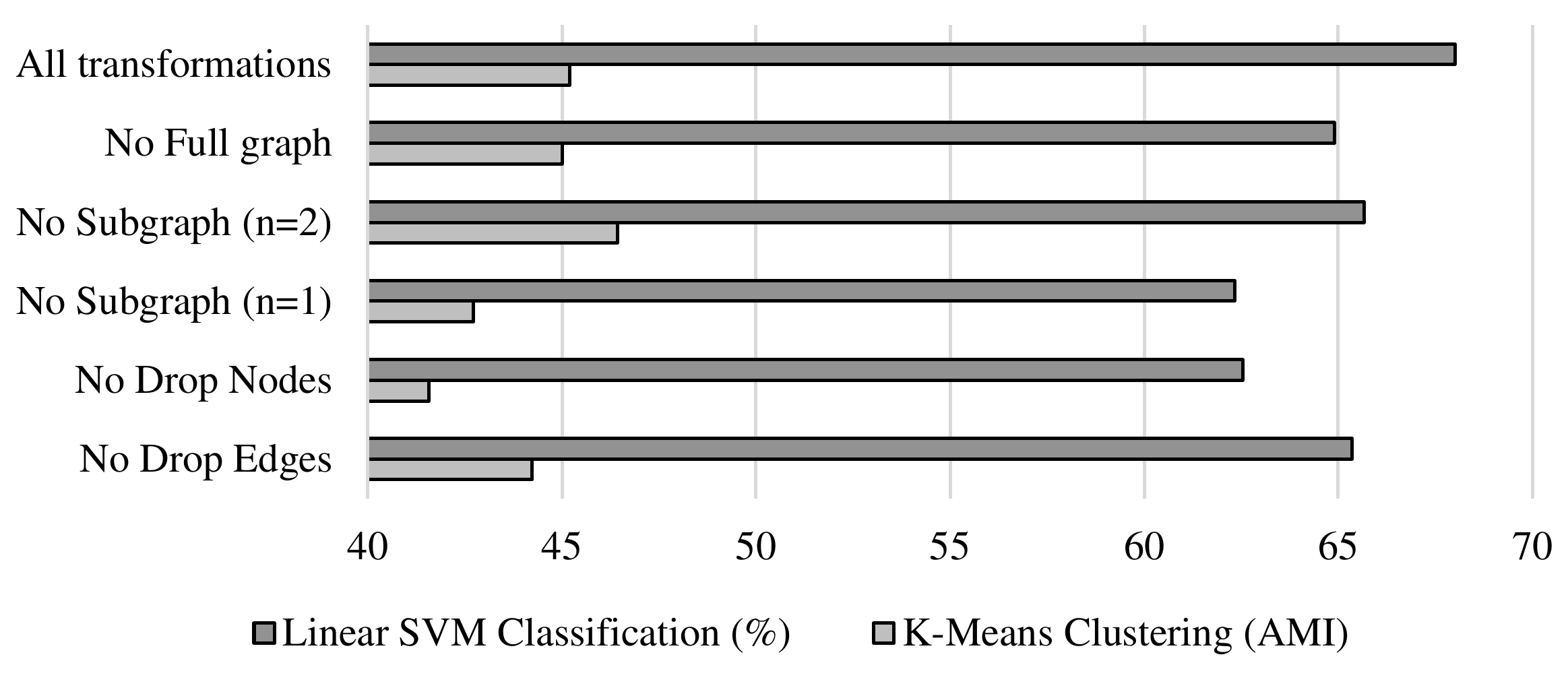}
    \caption{Ablation on the transformations used in contrastive learning.}
    \label{fig:ablation_retrieval}
\end{figure}
The clustering and linear SVM classification scores are computed as described in Section~\ref{ssec:clr} and reported in Figure~\ref{fig:ablation_retrieval}.
It is apparent from the results that using all the proposed transformations together is generally beneficial and improves the shape embeddings.
It is important to note that the transformations may have to be tuned for practical use cases based on the dataset and potential downstream tasks.
For example, the right number of hops used to define the subgraphs may vary based on the complexity of the B-reps in the dataset.
We did not explore this in our experiments, and directly applied the method that gave best results in the SolidLetters dataset on the ABC dataset.

\subsubsection{Shape retrieval}
\begin{figure*}
    \centering
    \includegraphics[height=0.94\textheight,keepaspectratio]{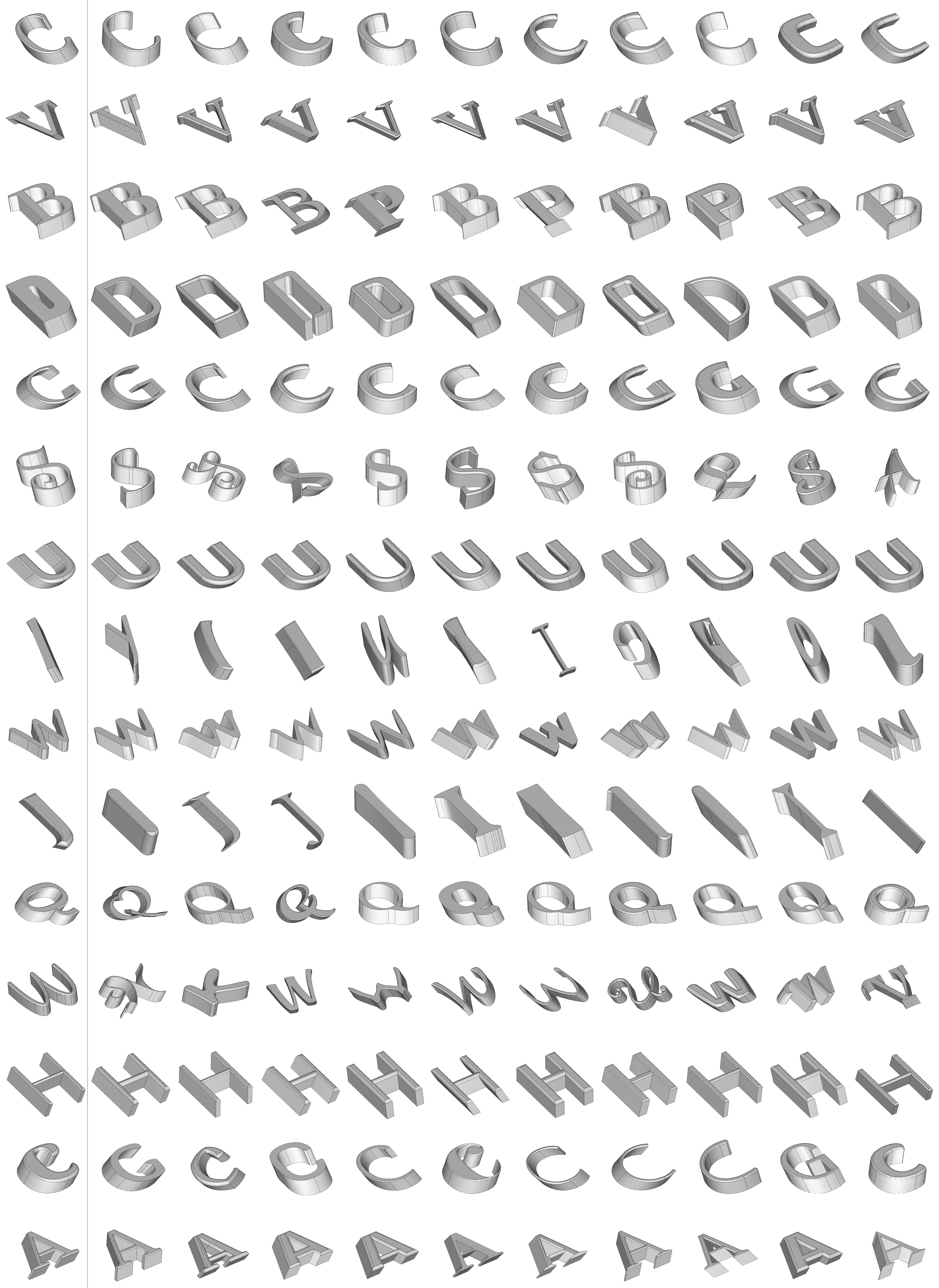}
    \caption{More self-supervised shape retrieval results on SolidLetters. Column 1: Query, Columns 2--11: Retrieved results sorted left to right by distance in latent space.}
    \label{fig:more_solidmnist_retrievals}
\end{figure*}

\begin{figure*}
    \centering
    \includegraphics[height=0.94\textheight,keepaspectratio]{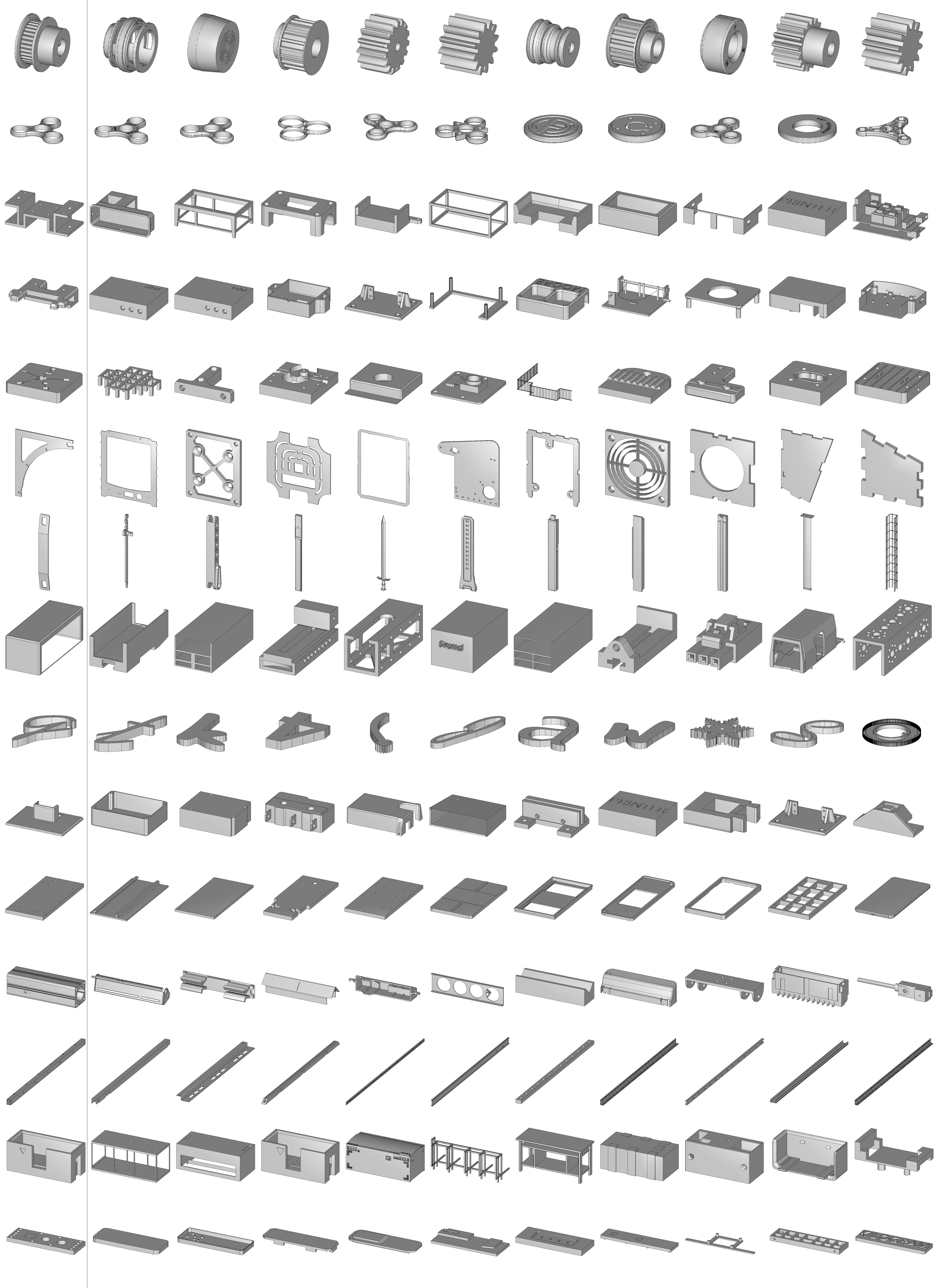}
    \caption{More self-supervised shape retrieval results on ABC. Column 1: Query, Columns 2--11: Retrieved results sorted left to right by distance in latent space.}
    \label{fig:more_abc_retrievals}
\end{figure*}

Here we share more qualitative results for shape retrieval on the SolidLetters and ABC datasets with k-nearest search in the latent space generated by our contrastive learning method in Figure~\ref{fig:more_solidmnist_retrievals} and Figure~\ref{fig:more_abc_retrievals}.

\end{document}